\documentclass[10pt,twocolumn]{article}

\usepackage{jfrExamplee}
\usepackage{graphics, dblfloatfix} 
\usepackage{epsfig} 
\usepackage{mathptmx} 
\usepackage{times} 
\usepackage{amsmath} 
\usepackage{amssymb}  
\usepackage{textcomp}
\usepackage{colortbl}
\usepackage{hhline}
\usepackage{subfigure}
\usepackage{bm}
\usepackage{booktabs}
\usepackage{multirow}
\usepackage[table,xcdraw]{xcolor}
\usepackage{caption} 
\usepackage[hyphens]{url}
\usepackage{hyperref}
\usepackage{titletoc}

\usepackage[left=2cm,right=2cm,top=2cm,bottom=2cm]{geometry}

\graphicspath{{figs/}}

\renewcommand{\vec}[1]{\mathbf{#1}}

\usepackage{setspace}

\title{A Survey of Deep Learning Techniques for Autonomous Driving}

\author{
Sorin Grigorescu\thanks{The authors are with \textit{Elektrobit Automotive} and the \textit{Robotics, Vision and Control Laboratory} (ROVIS Lab) at the Department of Automation and Information Technology, Transilvania University of Brasov, 500036 Romania. E-mail: (see \url{http://rovislab.com/sorin_grigorescu.html}).} \\
Artificial Intelligence,\\
Elektrobit Automotive.\\
Robotics, Vision and Control Lab,\\
Transilvania University of Brasov.\\
Brasov, Romania \\
\texttt{Sorin.Grigorescu@elektrobit.com} \\
\And
Bogdan Trasnea\\
Artificial Intelligence,\\
Elektrobit Automotive.\\
Robotics, Vision and Control Lab,\\
Transilvania University of Brasov.\\
Brasov, Romania \\
\texttt{Bogdan.Trasnea@elektrobit.com} \\
\AND
Tiberiu Cocias\\
Artificial Intelligence,\\
Elektrobit Automotive.\\
Robotics, Vision and Control Lab,\\
Transilvania University of Brasov.\\
Brasov, Romania \\
\texttt{Tiberiu.Cocias@elektrobit.com} \\
\And
Gigel Macesanu\\
Artificial Intelligence,\\
Elektrobit Automotive.\\
Robotics, Vision and Control Lab,\\
Transilvania University of Brasov.\\
Brasov, Romania \\
\texttt{Gigel.Macesanu@elektrobit.com} \\
}

\onecolumn

\begin{document}


\maketitle

\onehalfspacing
\begin{abstract}

The last decade witnessed increasingly rapid progress in self-driving vehicle technology, mainly backed up by advances in the area of deep learning and artificial intelligence. The objective of this paper is to survey the current state-of-the-art on deep learning technologies used in autonomous driving. We start by presenting AI-based self-driving architectures, convolutional and recurrent neural networks, as well as the deep reinforcement learning paradigm. These methodologies form a base for the surveyed driving scene perception, path planning, behavior arbitration and motion control algorithms. We investigate both the modular perception-planning-action pipeline, where each module is built using deep learning methods, as well as End2End systems, which directly map sensory information to steering commands. Additionally, we tackle current challenges encountered in designing AI architectures for autonomous driving, such as their safety, training data sources and computational hardware. The comparison presented in this survey helps to gain insight into the strengths and limitations of deep learning and AI approaches for autonomous driving and assist with design choices.\footnote{The articles referenced in this survey can be accessed at the web-page accompanying this paper, available at \url{http://rovislab.com/survey_DL_AD.html}}

\end{abstract}

\setlength{\columnsep}{0.8cm} 
\setlength{\parindent}{1em}
\setlength{\parskip}{0em}
\renewcommand{\baselinestretch}{1.5}

\newpage
\doublespacing
\tableofcontents

\newpage
\twocolumn
\singlespacing

\section{Introduction}
\label{sec:introduction}

Over the course of the last decade, Deep Learning and Artificial Intelligence (AI) became the main technologies behind many breakthroughs in computer vision~\cite{Alex_Krizhevsky_NIPS2012}, robotics~\cite{OpenAI_Learning_Dexterous} and Natural Language Processing (NLP)~\cite{Goldberg17NLP}. They also have a major impact in the autonomous driving revolution seen today both in academia and industry. Autonomous Vehicles (AVs) and self-driving cars began to migrate from laboratory development and testing conditions to driving on public roads. Their deployment in our environmental landscape offers a decrease in road accidents and traffic congestions, as well as an improvement of our mobility in overcrowded cities. The title of "self-driving" may seem self-evident, but there are actually five SAE Levels used to define autonomous driving. The SAE J3016 standard~\cite{SAE_2014} introduces a scale from 0 to 5 for grading vehicle automation. Lower SAE Levels feature basic driver assistance, whilst higher SAE Levels move towards vehicles requiring no human interaction whatsoever. Cars in the level 5 category require no human input and typically will not even feature steering wheels or foot pedals.

Although most driving scenarios can be relatively simply solved with classical perception, path planning and motion control methods, the remaining unsolved scenarios are corner cases in which traditional methods fail.



One of the first autonomous cars was developed by Ernst Dickmanns~\cite{Dickmanns1988} in the 1980s. This paved the way for new research projects, such as PROMETHEUS, which aimed to develop a fully functional autonomous car. In 1994, the VaMP driverless car managed to drive $1,600 km$, out of which $95\%$ were driven autonomously. Similarly, in 1995, CMU NAVLAB demonstrated autonomous driving on $6,000 km$, with $98\%$ driven autonomously. Another important milestone in autonomous driving were the DARPA Grand Challenges in 2004 and 2005, as well as the DARPA Urban Challenge in 2007. The goal was for a driverless car to navigate an off-road course as fast as possible, without human intervention. In 2004, none of the 15 vehicles completed the race. Stanley, the winner of the 2005 race, leveraged Machine Learning techniques for navigating the unstructured environment. This was a turning point in self-driving cars development, acknowledging Machine Learning and AI as central components of autonomous driving. The turning point is also notable in this survey paper, since the majority of the surveyed work is dated after 2005.

In this survey, we review the different artificial intelligence and deep learning technologies used in autonomous driving, and provide a survey on state-of-the-art deep learning and AI methods applied to self-driving cars. We also dedicate complete sections on tackling safety aspects, the challenge of training data sources and the required computational hardware.

\bigskip

\section{Deep Learning based Decision-Making Architectures for Self-Driving Cars}

\begin{figure*}
	\centering
	\begin{center}
		\includegraphics[scale=1.05]{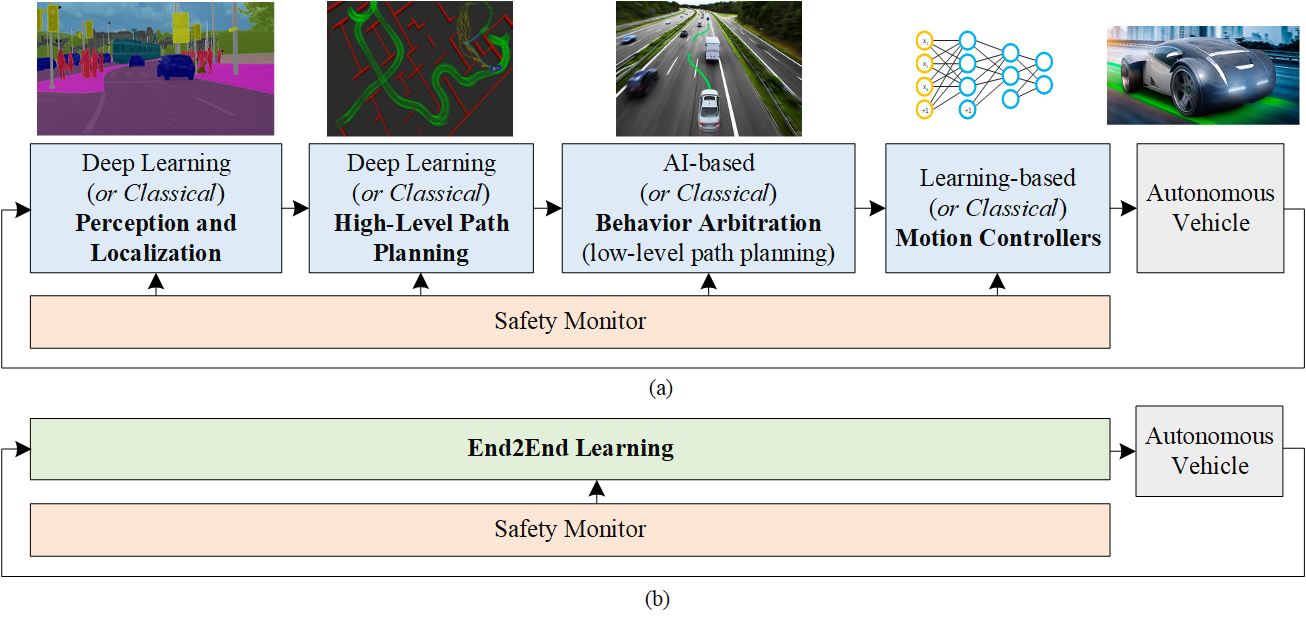}
		\caption{\textbf{Deep Learning based self-driving car}. The architecture can be implemented either as a sequential perception-planing-action pipeline (a), or as an End2End system (b). In the sequential pipeline case, the components can be designed either using AI and deep learning methodologies, or based on classical non-learning approaches. End2End learning systems are mainly based on deep learning methods. A safety monitor is usually designed to ensure the safety of each module.}
        \label{fig:ROB-19-0022_standard_had_system_fig1}
	\end{center}
\end{figure*}

Self-driving cars are autonomous decision-making systems that process streams of observations coming from different on-board sources, such as cameras, radars, LiDARs, ultrasonic sensors, GPS units and/or inertial sensors. These observations are used by the car's computer to make driving decisions. The basic block diagrams of an AI powered autonomous car are shown in Fig.~\ref{fig:ROB-19-0022_standard_had_system_fig1}. The driving decisions are computed either in a modular perception-planning-action pipeline (Fig.~\ref{fig:ROB-19-0022_standard_had_system_fig1}(a)), or in an End2End learning fashion (Fig.~\ref{fig:ROB-19-0022_standard_had_system_fig1}(b)), where sensory information is directly mapped to control outputs. The components of the modular pipeline can be designed either based on AI and deep learning methodologies, or using classical non-learning approaches. Various permutations of learning and non-learning based components are possible (e.g. a deep learning based object detector provides input to a classical A-star path planning algorithm). A safety monitor is designed to assure the safety of each module.

The modular pipeline in Fig.~\ref{fig:ROB-19-0022_standard_had_system_fig1}(a) is hierarchically decomposed into four components which can be designed using either deep learning and AI approaches, or classical methods. These components are:

\begin{itemize}
	\item \textit{Perception and Localization},
	\item \textit{High-Level Path Planning},
	\item \textit{Behavior Arbitration, or low-level path planning},
	\item \textit{Motion Controllers}.
\end{itemize}

Based on these four high-level components, we have grouped together relevant deep learning papers describing methods developed for autonomous driving systems. Additional to the reviewed algorithms, we have also grouped relevant articles covering the \textit{safety}, \textit{data sources} and \textit{hardware} aspects encountered when designing deep learning modules for self-driving cars.

Given a route planned through the road network, the first task of an autonomous car is to understand and localize itself in the surrounding environment. Based on this representation, a continuous path is planned and the future actions of the car are determined by the behavior arbitration system. Finally, a motion control system reactively corrects errors generated in the execution of the planned motion. A review of classical non-AI design methodologies for these four components can be found in~\cite{PadenCYYF16}.

Following, we will give an introduction of deep learning and AI technologies used in autonomous driving, as well as surveying different methodologies used to design the hierarchical decision making process described above. Additionally, we provide an overview of End2End learning systems used to encode the hierarchical process into a single deep learning architecture which directly maps sensory observations to control outputs.

\section{Overview of Deep Learning Technologies}

In this section, we describe the basis of deep learning technologies used in autonomous vehicles and comment on the capabilities of each paradigm. We focus on \textit{Convolutional Neural Networks} (CNN), \textit{Recurrent Neural Networks} (RNN) and \textit{Deep Reinforcement Learning} (DRL), which are the most common deep learning methodologies applied to autonomous driving.


Throughout the survey, we use the following notations to describe time dependent sequences. The value of a variable is defined either for a single discrete time step $t$, written as superscript $<t>$, or as a discrete sequence defined in the $<t, t+k>$ time interval, where $k$ denotes the length of the sequence. For example, the value of a state variable $\vec{z}$ is defined either at discrete time $t$, as $\vec{z}^{<t>}$, or within a sequence interval $\vec{z}^{<t, t+k>}$. Vectors and matrices are indicated by bold symbols.

\subsection{Deep Convolutional Neural Networks}

\textit{Convolutional Neural Networks} (CNN) are mainly used for processing spatial information, such as images, and can be viewed as image features extractors and universal non-linear function approximators~\cite{LeCun1998}, \cite{Bengio2013}. Before the rise of deep learning, computer vision systems used to be implemented based on handcrafted features, such as HAAR~\cite{ViolaJ01}, Local Binary Patterns (LBP)~\cite{Lbp1996}, or Histograms of Oriented Gradients (HoG)~\cite{Dalal05histogramsof}. In comparison to these traditional handcrafted features, convolutional neural networks are able to automatically learn a representation of the feature space encoded in the training set.

CNNs can be loosely understood as very approximate analogies to different parts of the mammalian visual cortex~\cite{Hubel1963}. An image formed on the retina is sent to the visual cortex through the thalamus. Each brain hemisphere has its own visual cortex. The visual information is received by the visual cortex in a crossed manner: the left visual cortex receives information from the right eye, while the right visual cortex is fed with visual data from the left eye. The information is processed according to the dual flux theory~\cite{Goodale1992}, which states that the visual flow follows two main fluxes: a \textit{ventral flux}, responsible for visual identification and object recognition, and a \textit{dorsal flux} used for establishing spatial relations between objects. A CNN mimics the functioning of the ventral flux, in which different areas of the brain are sensible to specific features in the visual field. The earlier brain cells in the visual cortex are activated by sharp transitions in the visual field of view, in the same way in which an edge detector highlights sharp transitions between the neighboring pixels in an image. These edges are further used in the brain to approximate object parts and finally to estimate abstract representations of objects.

An CNN is parametrized by its weights vector $\theta = [\vec{W}, \vec{b}]$, where $\vec{W}$ is the set of weights governing the inter-neural connections and $\vec{b}$ is the set of neuron bias values. The set of weights $\vec{W}$ is organized as image filters, with coefficients learned during training. Convolutional layers within a CNN exploit local spatial correlations of image pixels to learn translation-invariant convolution filters, which capture discriminant image features.

Consider a multichannel signal representation $\vec{M}_k$ in layer $k$, which is a channel-wise integration of signal representations $\vec{M}_{k,c}$, where $c \in \mathbb{N}$. A signal representation can be generated in layer $k+1$ as:

\begin{equation}
	\vec{M}_{k+1, l} = \varphi (\vec{M}_k * \vec{w}_{k,l} + \vec{b}_{k,l}),
\end{equation}

\noindent where $\vec{w}_{k,l} \in \vec{W}$ is a convolutional filter with the same number of channels as $\vec{M}_k$, $\vec{b}_{k,l} \in \vec{b}$ represents the bias, $l$ is a channel index and $*$ denotes the convolution operation. $\varphi (\cdot)$ is an activation function applied to each pixel in the input signal. Typically, the Rectified Linear Unit (ReLU) is the most commonly used activation function in computer vision applications~\cite{Alex_Krizhevsky_NIPS2012}. The final layer of a CNN is usually a fully-connected layer which acts as an object discriminator on a high-level abstract representation of objects.

In a supervised manner, the response $R(\cdot; \theta)$ of a CNN can be trained using a training database $\mathcal{D} = [(\vec{x}_1, y_1), ..., (\vec{x}_m, y_m)]$, where $\vec{x}_i$ is a data sample, $y_i$ is the corresponding label and $m$ is the number of training examples. The optimal network parameters can be calculated using \textit{Maximum Likelihood Estimation} (MLE). For the clarity of explanation, we take as example the simple least-squares error function, which can be used to drive the MLE process when training regression estimators:

\begin{equation}
	\vec{\hat{\theta}} = \arg \max_{\theta} \mathcal{L} (\theta; \mathcal{D}) = \arg \min_{\theta} \sum^{m}_{i=1} (R(\vec{x}_i; \theta) - y_i)^2.
	\label{eq:mle_cnn}
\end{equation}

\noindent For classification purposes, the least-squares error is usually replaced by the cross-entropy, or the negative log-likelihood loss functions. The optimization problem in Eq.~\ref{eq:mle_cnn} is typically solved with Stochastic Gradient Descent (SGD) and the backpropagation algorithm for gradient estimation~\cite{Rumelhart1986}. In practice, different variants of SGD are used, such as Adam~\cite{Adam_Optimizer_2015} or AdaGrad~\cite{AdaGrad_Optimizer_2011}.

\subsection{Recurrent Neural Networks}

\begin{figure}
	\centering
	\begin{center}
		\includegraphics[scale=1.0]{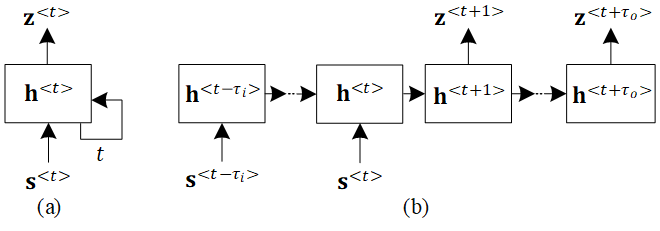}
		\caption{\textbf{A folded (a) and unfolded (b) over time, many-to-many Recurrent Neural Network}. Over time $t$, both the input $\vec{s}^{<t-\tau_i, t>}$ and output $\vec{z}^{<t+1, t+\tau_o>}$ sequences share the same weights $\vec{h}^{<\cdot>}$. The architecture is also referred to as a \textit{sequence-to-sequence model}.}
        \label{ROB-19-0022_rnn_block_diagram_fig2}
	\end{center}
	\vspace{-1.9em}
\end{figure}

Among deep learning techniques, \textit{Recurrent Neural Networks} (RNN) are especially good in processing temporal sequence data, such as text, or video streams. Different from conventional neural networks, a RNN contains a time dependent feedback loop in its memory cell. Given a time dependent input sequence $[\vec{s}^{<t-\tau_i>}, ..., \vec{s}^{<t>}]$ and an output sequence $[\vec{z}^{<t+1>}, ..., \vec{z}^{<t+\tau_o>}]$, a RNN can be "unfolded" $\tau_i + \tau_o$ times to generate a loop-less network architecture matching the input length, as illustrated in Fig.~\ref{ROB-19-0022_rnn_block_diagram_fig2}. $t$ represents a temporal index, while $\tau_i$ and $\tau_o$ are the lengths of the input and output sequences, respectively. Such neural networks are also encountered under the name of \textit{sequence-to-sequence models}. An unfolded network has $\tau_i + \tau_o + 1$ identical layers, that is, each layer shares the same learned weights. Once unfolded, a RNN can be trained using the backpropagation through time algorithm. When compared to a conventional neural network, the only difference is that the learned weights in each unfolded copy of the network are averaged, thus enabling the network to shared the same weights over time.

The main challenge in using basic RNNs is the vanishing gradient encountered during training. The gradient signal can end up being multiplied a large number of times, as many as the number of time steps. Hence, a traditional RNN is not suitable for capturing long-term dependencies in sequence data. If a network is very deep, or processes long sequences, the gradient of the network's output would have a hard time in propagating back to affect the weights of the earlier layers. Under gradient vanishing, the weights of the network will not be effectively updated, ending up with very small weight values.

Long Short-Term Memory (LSTM)~\cite{hochreiter1997long} networks are non-linear function approximators for estimating temporal dependencies in sequence data. As opposed to traditional recurrent neural networks, LSTMs solve the vanishing gradient problem by incorporating three gates, which control the input, output and memory state.


Recurrent layers exploit temporal correlations of sequence data to learn time dependent neural structures. Consider the memory state $\vec{c}^{<t-1>}$ and the output state $\vec{h}^{<t-1>}$ in an LSTM network, sampled at time step $t-1$, as well as the input data $\vec{s}^{<t>}$ at time $t$. The opening or closing of a gate is controlled by a sigmoid function $\sigma(\cdot)$ of the current input signal $\vec{s}^{<t>}$ and the output signal of the last time point $\vec{h}^{<t-1>}$, as follows:

\begin{equation}
	\Gamma_u^{<t>} = \sigma ( \vec{W}_u \vec{s}^{<t>} + \vec{U}_u \vec{h}^{<t-1>} + \vec{b}_u),
	\label{eq:update_gate}
\end{equation}

\begin{equation}
	\Gamma_f^{<t>} = \sigma ( \vec{W}_f \vec{s}^{<t>} + \vec{U}_f \vec{h}^{<t-1>} + \vec{b}_f),
	\label{eq:forget_gate}
\end{equation}

\begin{equation}
	\Gamma_o^{<t>} = \sigma ( \vec{W}_o \vec{s}^{<t>} + \vec{U}_o \vec{h}^{<t-1>} + \vec{b}_o),
	\label{eq:output_gate}
\end{equation}

\noindent where $\Gamma_u^{<t>}$, $\Gamma_f^{<t>}$ and $\Gamma_o^{<t>}$ are gate functions of the input gate, forget gate and output gate, respectively. Given current observation, the memory state $\vec{c}^{<t>}$ will be updated as:

\begin{equation}
	\vec{c}^{<t>} = \Gamma_u^{<t>} * \tanh ( \vec{W}_c \vec{s}^{<t>} + \vec{U}_c \vec{h}^{<t-1>} + \vec{b}_c) + \Gamma_f * \vec{c}^{<t-1>},
	\label{eq:memory_cell_update}
\end{equation}

\noindent The new network output $\vec{h}^{<t>}$ is computed as:

\begin{equation}
	\vec{h}^{<t>} = \Gamma_o^{<t>} * \tanh ( \vec{c}^{<t>}).
	\label{eq:memory_cell_update}
\end{equation}

An LSTM network $Q$ is parametrized by $\theta = [\vec{W}_i, \vec{U}_i, \vec{b}_i]$, where $\vec{W}_i$ represents the weights of the network's gates and memory cell multiplied with the input state, $\vec{U}_i$ are the weights governing the activations and $\vec{b}_i$ denotes the set of neuron bias values. $*$ symbolizes element-wise multiplication.

In a supervised learning setup, given a set of training sequences $\mathcal{D} = [(\vec{s}^{<t-\tau_i, t>}_1, \vec{z}^{<t+1, t+\tau_o>}_1), ..., (\vec{s}^{<t-\tau_i, t>}_q, \vec{z}^{<t+1, t+\tau_o>}_q)]$, that is, $q$ independent pairs of observed sequences with assignments $\vec{z}^{<t, t+\tau_o>}$, one can train the response of an LSTM network $Q(\cdot; \theta)$ using Maximum Likelihood Estimation:

\begin{equation}
	\begin{split}
		\hat{\theta} & = \arg \max_{\theta} \mathcal{L} (\theta; \mathcal{D}) \\
		& = \arg \min_{\theta} \sum_{i=1}^m l_i (Q (\vec{s}^{<t-\tau_i, t>}_i; \theta), \vec{z}^{<t+1, t+\tau_o>}_i), \\
		& = \arg \min_{\theta} \sum_{i=1}^m \sum_{t=1}^{\tau_o} l_i^{<t>} (Q^{<t>} (\vec{s}^{<t-\tau_i, t>}_i; \theta), \vec{z}^{<t>}_i),
	\end{split}
	\label{eq:nn_mle_training}
\end{equation}

\noindent where an input sequence of observations $\vec{s}^{<t-\tau_i, t>} = [\vec{s}^{<t-\tau_i>}, ..., \vec{s}^{<t-1>}, \vec{s}^{<t>}]$ is composed of $\tau_i$ consecutive data samples, $l(\cdot,\cdot)$ is the logistic regression loss function and $t$ represents a temporal index.

In recurrent neural networks terminology, the optimization procedure in Eq.~\ref{eq:nn_mle_training} is typically used for training "many-to-many" RNN architectures, such as the one in Fig.~\ref{ROB-19-0022_rnn_block_diagram_fig2}, where the input and output states are represented by temporal sequences of $\tau_i$ and $\tau_o$ data instances, respectively. This optimization problem is commonly solved using gradient based methods, like Stochastic Gradient Descent (SGD), together with the backpropagation through time algorithm for calculating the network's gradients.

\subsection{Deep Reinforcement Learning}

In the following, we review the \textit{Deep Reinforcement Learning} (DRL) concept as an autonomous driving task, using the \textit{Partially Observable Markov Decision Process} (POMDP) formalism.

In a POMDP, an agent, which in our case is the self-driving car, senses the environment with observation $\vec{I}^{<t>}$, performs an action $a^{<t>}$ in state $\vec{s}^{<t>}$, interacts with its environment through a received reward $R^{<t+1>}$, and transits to the next state $\vec{s}^{<t+1>}$ following a transition function $T_{\vec{s}^{<t>}, a^{<t>}}^{\vec{s}^{<t+1>}}$.

In RL based autonomous driving, the task is to learn an optimal driving policy for navigating from state $\vec{s}^{<t>}_{start}$ to a destination state $\vec{s}^{<t+k>}_{dest}$, given an observation $\vec{I}^{<t>}$ at time $t$ and the system's state $\vec{s}^{<t>}$. $\vec{I}^{<t>}$ represents the observed environment, while $k$ is the number of time steps required for reaching the destination state $\vec{s}^{<t+k>}_{dest}$.

In reinforcement learning terminology, the above problem can be modeled as a POMDP $M := (I, S, A, T, R, \gamma)$, where:

\begin{itemize}
	\item $I$ is the set of observations, with $\vec{I}^{<t>} \in I$ defined as an observation of the environment at time $t$.
	
	\item $S$ represents a finite set of states, $\vec{s}^{<t>} \in S$ being the state of the agent at time $t$, commonly defined as the vehicle's position, heading and velocity.
	
	\item $A$ represents a finite set of actions allowing the agent to navigate through the environment defined by $\vec{I}^{<t>}$, where $a^{<t>} \in A$ is the action performed by the agent at time $t$.
	
	\item $T: S \times A \times S \rightarrow [0, 1]$ is a stochastic transition function, where $T_{\vec{s}^{<t>}, a^{<t>}}^{\vec{s}^{<t+1>}}$ describes the probability of arriving in state $\vec{s}^{<t+1>}$, after performing action $a^{<t>}$ in state $\vec{s}^{<t>}$.
	
	\item $R: S \times A \times S \rightarrow \mathbb{R}$ is a scalar reward function which controls the estimation of $a$, where $R_{\vec{s}^{<t>}, a^{<t>}}^{\vec{s}^{<t+1>}} \in \mathbb{R}$. For a state transition $\vec{s}^{<t>} \rightarrow \vec{s}^{<t+1>}$ at time $t$, we define a scalar reward function $R_{\vec{s}^{<t>}, a^{<t>}}^{\vec{s}^{<t+1>}}$ which quantifies how well did the agent perform in reaching the next state.
	
	\item $\gamma$ is the discount factor controlling the importance of future versus immediate rewards.
\end{itemize}

Considering the proposed reward function and an arbitrary state trajectory $[\vec{s}^{<0>}, \vec{s}^{<1>}, ..., \vec{s}^{<k>}]$ in observation space, at any time $\hat{t} \in [0, 1, ..., k]$, the associated cumulative future discounted reward is defined as:

\begin{equation}
	R^{<\hat{t}>} = \sum^{k}_{t=\hat{t}} \gamma^{<t-\hat{t}>} r^{<t>},
	\label{eq:cumulative_reward}
\end{equation}

\noindent where the immediate reward at time $t$ is given by $r^{<t>}$. In RL theory, the statement in Eq.~\ref{eq:cumulative_reward} is known as a finite horizon learning episode of sequence length $k$~\cite{Sutton_RL}.

The objective in RL is to find the desired trajectory policy that maximizes the associated cumulative future reward. We define the optimal action-value function $Q^*(\cdot, \cdot)$ which estimates the maximal future discounted reward when starting in state $\vec{s}^{<t>}$ and performing actions $[a^{<t>}, ..., a^{<t+k>}]$:

\begin{equation}
	Q^* (\vec{s}, a) = \underset{\pi}{\max} \mathbb{E} \text{ } [R^{<\hat{t}>} | \vec{s}^{<\hat{t}>} = \vec{s}, \text{ } a^{<\hat{t}>} = a, \text{ } \pi],
	\label{eq:optimal_action_value_function}
\end{equation}

\noindent where $\pi$ is an action policy, viewed as a probability density function over a set of possible actions that can take place in a given state. The optimal action-value function $Q^*(\cdot, \cdot)$ maps a given state to the optimal action policy of the agent in any state:

\begin{equation}
	\forall \vec{s} \in S: \pi^* (\vec{s}) = \underset{a \in A}{\arg\max} Q^* (\vec{s}, a).
\end{equation}

The optimal action-value function $Q^*$ satisfies the Bellman optimality equation~\cite{Bellman}, which is a recursive formulation of Eq.~\ref{eq:optimal_action_value_function}:

\begin{equation}
\begin{aligned}
		Q^* (\vec{s}, a) = \sum_{\vec{s}} T_{\vec{s}, a}^{\vec{s}'} \left( R_{\vec{s}, a}^{\vec{s}'} + \gamma \cdot \underset{a'}{\max} Q^* (\vec{s}', a') \right) \\
		= \mathbb{E}_{a'} \left( R_{\vec{s}, a}^{\vec{s}'} + \gamma \cdot \underset{a'}{\max} Q^* (\vec{s}', a') \right),
\end{aligned}
	\label{eq:bellman_optimality_equation}
\end{equation}

\noindent where $\vec{s}'$ represents a possible state visited after $\vec{s} = \vec{s}^{<t>}$ and $a'$ is the corresponding action policy. The model-based policy iteration algorithm was introduced in~\cite{Sutton_RL}, based on the proof that the Bellman equation is a contraction mapping~\cite{Watkins_Q_Learning} when written as an operator $\nu$:

\begin{equation}
	\forall Q, \lim_{n \rightarrow \infty} \nu^{(n)} (Q) = Q^*.
\end{equation}

However, the standard reinforcement learning method described above is not feasible in high dimensional state spaces. In autonomous driving applications, the observation space is mainly composed of sensory information made up of images, radar, LiDAR, etc. Instead of the traditional approach, a non-linear parametrization of $Q^*$ can be encoded in the layers of a deep neural network. In literature, such a non-linear approximator is called a Deep Q-Network (DQN)~\cite{mnih2015humanlevel} and is used for estimating the approximate action-value function:

\begin{equation}
	Q (\vec{s}^{<t>}, a^{<t>}; \Theta) \approx Q^* (\vec{s}^{<t>}, a^{<t>}),
\end{equation}

\noindent where $\Theta$ represents the parameters of the Deep Q-Network.

By taking into account the Bellman optimality equation~\ref{eq:bellman_optimality_equation}, it is possible to train a deep Q-network in a reinforcement learning manner through the minimization of the mean squared error. The optimal expected Q value can be estimated within a training iteration $i$ based on a set of reference parameters $\bar{\Theta}_i$ calculated in a previous iteration $i'$:

\begin{equation}
	y = R_{\vec{s}, a}^{\vec{s}'} + \gamma \cdot \underset{a'}{\max} Q(\vec{s}', a'; \bar{\Theta}_i),
\end{equation}

\noindent where $\bar{\Theta}_i := \Theta_{i'}$. The new estimated network parameters at training step $i$ are evaluated using the following squared error function:

\begin{equation}
	\nabla J_{\hat{\Theta}_i} = \underset{\Theta_i}{\min} \text{ } \mathbb{E}_{\vec{s}, y, r, \vec{s}'} \left[ \left( y - Q(\vec{s}, a; \Theta_i) \right)^2 \right],
	\label{eq:rl_setup}
\end{equation}

\noindent where $r = R_{\vec{s}, a}^{\vec{s}'}$. Based on~\ref{eq:rl_setup}, the maximum likelihood estimation function from Eq.~\ref{eq:nn_mle_training} can be applied for calculating the weights of the deep Q-network. The gradient is approximated with random samples and the backpropagation algorithm, which uses stochastic gradient descent for training:

\begin{equation}
	\nabla_{\Theta_i} = \mathbb{E}_{\vec{s}, a, r, \vec{s}'} \left[ \left( y - Q(\vec{s}, a; \Theta_i) \right) \nabla_{\Theta_i} \left( Q(\vec{s}, a; \Theta_i) \right) \right].
	\label{eq:gradient_descent}
\end{equation}


The deep reinforcement learning community has made several independent improvements to the original DQN algorithm~\cite{mnih2015humanlevel}. A study on how to combine these improvements on deep reinforcement learning has been provided by DeepMind in~\cite{hessel2017rainbow}, where the combined algorithm, entitled \textit{Rainbow}, was able to outperform the independently competing methods. DeepMind~\cite{hessel2017rainbow} proposes six extensions to the base DQN, each addressing a distinct concern:

\begin{itemize}
	\item \textit{Double Q Learning} addresses the overestimation bias and decouples the selection of an action and its evaluation;
	\item \textit{Prioritized replay} samples more frequently from the data in which there is information to learn;
	\item \textit{Dueling Networks} aim at enhancing value based RL;
	\item \textit{Multi-step learning} is used for training speed improvement;
	\item \textit{Distributional RL} improves the target distribution in the Bellman equation;
	\item \textit{Noisy Nets} improve the ability of the network to ignore noisy inputs and allows state-conditional exploration.
\end{itemize}

All of the above complementary improvements have been tested on the Atari 2600 challenge. A good implementation of DQN regarding autonomous vehicles should start by combining the stated DQN extensions with respect to a desired performance. Given the advancements in deep reinforcement learning, the direct application of the algorithm still needs a training pipeline in which one should simulate and model the desired self-driving car's behavior.

The simulated environment state is not directly accessible to the agent. Instead, sensor readings provide clues about the true state of the environment. In order to decode the true environment state, it is not sufficient to map a single snapshot of sensors readings. The temporal information should also be included in the network's input, since the environment's state is modified over time. An example of DQN applied to autonomous vehicles in a simulator can be found in~\cite{Sallab2017}.

DQN has been developed to operate in discrete action spaces. In the case of an autonomous car, the discrete actions would translate to discrete commands, such as turn left, turn right, accelerate, or break. The DQN approach described above has been extended to continuous action spaces based on policy gradient estimation~\cite{Lillicrap2016ContinuousCW}. The method in~\cite{Lillicrap2016ContinuousCW} describes a model-free actor-critic algorithm able to learn different continuous control tasks directly from raw pixel inputs. A model-based solution for continuous Q-learning is proposed in~\cite{GuLilSutLev16}.

Although continuous control with DRL is possible, the most common strategy for DRL in autonomous driving is based on discrete control~\cite{Jaritz2018}. The main challenge here is the training, since the agent has to explore its environment, usually through learning from collisions. Such systems, trained solely on simulated data, tend to learn a biased version of the driving environment. A solution here is to use Imitation Learning methods, such as Inverse Reinforcement Learning (IRL)~\cite{Wulfmeier2016}, to learn from human driving demonstrations without needing to explore unsafe actions.




\section{Deep Learning for Driving Scene Perception and Localization}
\label{sec:perception}

The self-driving technology enables a vehicle to operate autonomously by perceiving the environment and instrumenting a responsive answer.
Following, we give an overview of the top methods used in driving scene understanding, considering camera based vs. LiDAR environment perception. We survey object detection and recognition, semantic segmentation and localization in autonomous driving, as well as scene understanding using occupancy maps. Surveys dedicated to autonomous vision and environment perception can be found in~\cite{Zhu2017OverviewOE} and~\cite{Janai2017}.

\subsection{Sensing Hardware: Camera vs. LiDAR Debate}
\label{sec:camera_vs_lidar}

Deep learning methods are particularly well suited for detecting and recognizing objects in 2D images and 3D point clouds acquired from video cameras and LiDAR (Light Detection and Ranging) devices, respectively.

In the autonomous driving community, 3D perception is mainly based on LiDAR sensors, which provide a direct 3D representation of the surrounding environment in the form of 3D point clouds. The performance of a LiDAR is measured in terms of field of view, range, resolution and rotation/frame rate. 3D sensors, such as Velodyne\textsuperscript{\textregistered}, usually have a $360^{\circ}$ horizontal field of view. In order to operate at high speeds, an autonomous vehicle requires a minimum of $200m$ range, allowing the vehicle to react to changes in road conditions in time. The 3D object detection precision is dictated by the resolution of the sensor, with most advanced LiDARs being able to provide a $3cm$ accuracy.

\begin{figure*}
	\centering
	\begin{center}
		\includegraphics[scale=1.05]{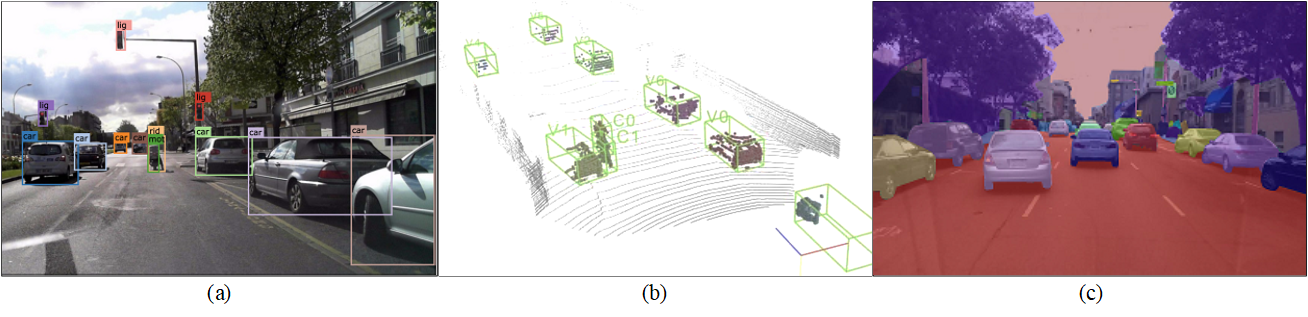}
		\caption{\textbf{Examples of scene perception results.} (a) 2D object detection in images. (b) 3D bounding box detector applied on LiDAR data. (c) Semantic segmentation results on images.}
        \label{ROB-19-0022_perception_fig3}
	\end{center}
\end{figure*}

Recent debate sparked around camera vs. LiDAR (Light Detection and Ranging) sensing technologies. Tesla\textsuperscript{\textregistered} and Waymo\textsuperscript{\textregistered}, two of the companies leading the development of self-driving technology~\cite{Sean_2018}, have different philosophies with respect to their main perception sensor, as well as regarding the targeted SAE level~\cite{SAE_2014}. Waymo\textsuperscript{\textregistered} is building their vehicles directly as Level 5 systems, with currently more than 10 million miles driven autonomously\footnote{\url{https://arstechnica.com/cars/2018/10/waymo-has-driven-10-million-miles-on-public-roads/}}. On the other hand, Tesla\textsuperscript{\textregistered} deploys its AutoPilot as an ADAS (Advanced Driver Assistance System) component, which customers can turn on or off at their convenience. The advantage of Tesla\textsuperscript{\textregistered} resides in its large training database, consisting of more than 1 billion driven miles\footnote{\url{https://electrek.co/2018/11/28/tesla-autopilot-1-billion-miles/}}. The database has been acquired by collecting data from customers-owned cars.

The main sensing technologies differ in both companies. Tesla\textsuperscript{\textregistered} tries to leverage on its camera systems, whereas Waymo's driving technology relies more on Lidar sensors\footnote{\url{https://www.theverge.com/transportation/2018/4/19/17204044/tesla-waymo-self-driving-car-data-simulation}}. The sensing approaches have advantages and disadvantages. LiDARs have high resolution and precise perception even in the dark, but are vulnerable to bad weather conditions (e.g. heavy rain)~\cite{7795918} and involve moving parts. In contrast, cameras are cost efficient, but lack depth perception and cannot work in the dark. Cameras are also sensitive to bad weather, if the weather conditions are obstructing the field of view.

Researchers at Cornell University tried to replicate LiDAR-like point clouds from visual depth estimation~\cite{wang2019pseudo}. An estimated depth map is reprojected into 3D space, with respect to the left sensor's coordinate of a stereo camera. The resulting point cloud is referred to as \textit{pseudo-LiDAR}. The pseudo-LiDAR data can be further fed to 3D deep learning processing methods, such as PointNet~\cite{Qi2017PointNet} or AVOD~\cite{Ku2018iros}. The success of image based 3D estimation is of high importance to the large scale deployment of autonomous cars, since the LiDAR is arguably one of the most expensive hardware component in a self-driving vehicle.

Apart from these sensing technologies, radar and ultrasonic sensors are used to enhance perception capabilities. For example, alongside three Lidar sensors, Waymo also makes use of five radars and eight cameras, while Tesla\textsuperscript{\textregistered} cars are equipped with eights cameras, 12 ultrasonic sensors and one forward-facing radar.

\subsection{Driving Scene Understanding}

An autonomous car should be able to detect traffic participants and drivable areas, particularly in urban areas where a wide variety of object appearances and occlusions may appear. Deep learning based perception, in particular Convolutional Neural Networks (CNNs), became the de-facto standard in object detection and recognition, obtaining remarkable results in competitions such as the ImageNet Large Scale Visual Recognition Challenge~\cite{ILSVRC15}.

Different neural networks architectures are used to detect objects as 2D regions of interest~\cite{redmon2016you}~\cite{law2018cornernet}~\cite{zhang2017single}~\cite{girshick2015fast}~\cite{iandola2016squeezenet}~\cite{dai2016r} or pixel-wise segmented areas in images~\cite{BadrinarayananK15}~\cite{zhao2017icnet}~\cite{Treml2016SpeedingUS}~\cite{He2017MaskR}, 3D bounding boxes in LiDAR point clouds~\cite{Qi2017PointNet}~\cite{Zhou2018VoxelNetEL}~\cite{Luo_2018_CVPR}, as well as 3D representations of objects in combined camera-LiDAR data~\cite{Qi2018FrustumPointNet}~\cite{Chen2017MultiView3D}~\cite{Ku2018iros}. Examples of scene perception results are illustrated in Fig.~\ref{ROB-19-0022_perception_fig3}. Being richer in information, image data is more suited for the object recognition task. However, the real-world 3D positions of the detected objects have to be estimated, since depth information is lost in the projection of the imaged scene onto the imaging sensor.

\subsubsection{Bounding-Box-Like Object Detectors}

The most popular architectures for 2D object detection in images are single stage and double stage detectors. Popular single stage detectors are "\textit{You Only Look Once}" (Yolo)~\cite{redmon2016you}~\cite{redmon2017yolo9000}~\cite{redmon2018yolov3}, the \textit{Single Shot multibox Detector} (SSD)~\cite{liu2016ssd}, CornerNet~\cite{law2018cornernet} and RefineNet~\cite{zhang2017single}. Double stage detectors, such as RCNN~\cite{Girshick_2014}, Faster-RCNN~\cite{ren2017faster}, or R-FCN~\cite{dai2016r}, split the object detection process into two parts: region of interest candidates proposals and bounding boxes classification. In general, single stage detectors do not provide the same performances as double stage detectors, but are significantly faster.

If in-vehicle computation resources are scarce, one can use detectors such as SqueezeNet~\cite{iandola2016squeezenet} or~\cite{li2018efficient}, which are optimized to run on embedded hardware. These detectors usually have a smaller neural network architecture, making it possible to detect objects using a reduced number of operations, at the cost of detection accuracy.

\begin{figure}
	\centering
	\begin{center}
		\includegraphics[scale=0.41]{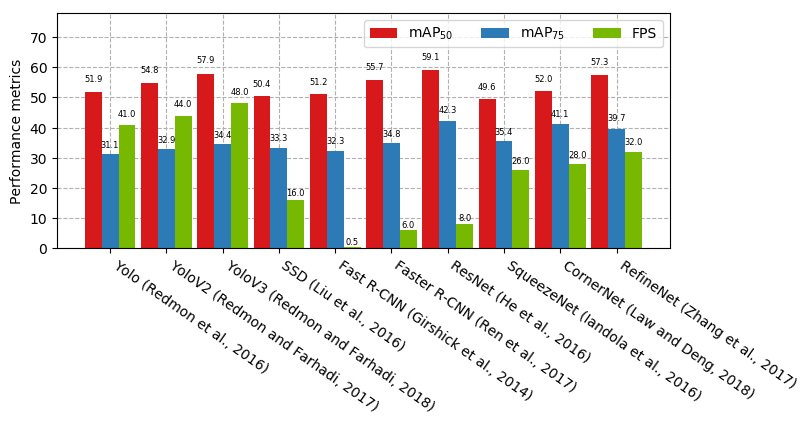}
		\caption{\textbf{Object detection and recognition performance comparison.} The evaluation has been performed on the Pascal VOC 2012 benchmarking database. The first four methods on the right represent \textit{single stage} detectors, while the remaining six are \textit{double stage} detectors. Due to their increased complexity, the runtime performance in Frames-per-Second (FPS) is lower for the case of double stage detectors.}
        \label{fig:ROB-19-0022_CNN_detectors_mAP_vs_FPS_fig4}
	\end{center}
	\vspace{-1.8em}
\end{figure}

A comparison between the object detectors described above is given in Figure~\ref{fig:ROB-19-0022_CNN_detectors_mAP_vs_FPS_fig4}, based on the Pascal VOC 2012 dataset and their measured mean Average Precision (mAP) with an Intersection over Union (IoU) value equal to $50$ and $75$, respectively.

A number of publications showcased object detection on raw 3D sensory data, as well as for combined video and LiDAR information. PointNet~\cite{Qi2017PointNet} and VoxelNet~\cite{Zhou2018VoxelNetEL} are designed to detect objects solely from 3D data, providing also the 3D positions of the objects. However, point clouds alone do not contain the rich visual information available in images. In order to overcome this, combined camera-LiDAR architectures are used, such as Frustum PointNet~\cite{Qi2018FrustumPointNet}, Multi-View 3D networks (MV3D)~\cite{Chen2017MultiView3D}, or RoarNet~\cite{Shin2018RoarNetAR}.

The main disadvantage in using a LiDAR in the sensory suite of a self-driving car is primarily its cost\footnote{\url{https://techcrunch.com/2019/03/06/waymo-to-start-selling-standalone-lidar-sensors/}}. A solution here would be to use neural network architectures such as AVOD (Aggregate View Object Detection)~\cite{Ku2018iros}, which leverage on LiDAR data only for training, while images are used during training and deployment. At deployment stage, AVOD is able to predict 3D bounding boxes of objects solely from image data. In such a system, a LiDAR sensor is necessary only for training data acquisition, much like the cars used today to gather road data for navigation maps.

\subsubsection{Semantic and Instance Segmentation}

Driving scene understanding can also be achieved using semantic segmentation, representing the categorical labeling of each pixel in an image. In the autonomous driving context, pixels can be marked with categorical labels representing drivable area, pedestrians, traffic participants, buildings, etc. It is one of the high-level tasks that paves the way towards complete scene understanding, being used in applications such as autonomous driving, indoor navigation, or virtual and augmented reality.

Semantic segmentation networks like SegNet~\cite{BadrinarayananK15}, ICNet~\cite{zhao2017icnet}, ENet~\cite{paszke2016enet}, AdapNet~\cite{Valada2017AdapNetAS}, or Mask R-CNN~\cite{He2017MaskR} are mainly encoder-decoder architectures with a pixel-wise classification layer. These are based on building blocks from some common network topologies, such as AlexNet \cite{Alex_Krizhevsky_NIPS2012}, VGG-16 \cite{simonyan2014very}, GoogLeNet \cite{szegedy2014going}, or ResNet \cite{he2016deep}. 

As in the case of bounding-box detectors, efforts have been made to improve the computation time of these systems on embedded targets. In \cite{Treml2016SpeedingUS} and~\cite{paszke2016enet}, the authors proposed approaches to speed up data processing and inference on embedded devices for autonomous driving. Both architectures are light networks providing similar results as SegNet, with a reduced computation cost.

The robustness objective for semantic segmentation was tackled for optimization in AdapNet~\cite{Valada2017AdapNetAS}. The model is capable of robust segmentation in various environments by adaptively learning features of expert networks based on scene conditions.

A combined bounding-box object detector and semantic segmentation result can be obtained using architectures such as Mask R-CNN~\cite{He2017MaskR}. The method extends the effectiveness of Faster-RCNN to instance segmentation by adding a branch for predicting an object mask in parallel with the existing branch for bounding box recognition.

\begin{figure}
	\centering
	\begin{center}
		\includegraphics[scale=0.55]{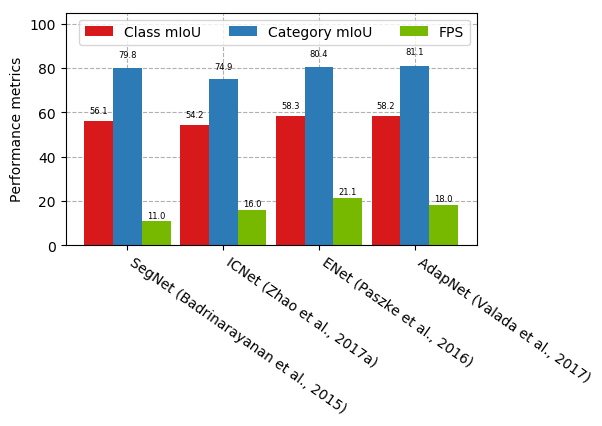}
		\caption{\textbf{Semantic segmentation performance comparison on the CityScapes dataset}~\cite{Cityscapes2018}. The input samples are $480px \times 320px$ images of driving scenes.}
        \label{fig:ROB-19-0022_Segmentation_mIoU_vs_FPS_fig5}
	\end{center}
	\vspace{-1.8em}
\end{figure}

Figure~\ref{fig:ROB-19-0022_Segmentation_mIoU_vs_FPS_fig5} shows tests results performed on four key semantic segmentation networks, based on the CityScapes dataset. The per-class mean Intersection over Union (mIoU) refers to multi-class segmentation, where each pixel is labeled as belonging to a specific object class, while per-category mIoU refers to foreground (object) - background (non-object) segmentation. The input samples have a size of $480px \times 320px$.

\subsubsection{Localization}

\textit{Localization} algorithms aim at calculating the pose (position and orientation) of the autonomous vehicle as it navigates. Although this can be achieved with systems such as GPS, in the followings we will focus on deep learning techniques for visual based localization.

Visual Localization, also known as \textit{Visual Odometry} (VO), is typically determined by matching keypoint landmarks in consecutive video frames. Given the current frame, these keypoints are used as input to a perspective-$n$-point mapping algorithm for computing the pose of the vehicle with respect to the previous frame. Deep learning can be used to improve the accuracy of VO by directly influencing the precision of the keypoints detector. In~\cite{BarnesICRA2018}, a deep neural network has been trained for learning keypoints distractors in monocular VO. The so-called learned ephemerality mask, acts a a rejection scheme for keypoints outliers which might decrease the vehicle localization's accuracy. The structure of the environment can be mapped incrementally with the computation of the camera pose. These methods belong to the area of \textit{Simultaneous Localization and Mapping} (SLAM). For a survey on classical SLAM techniques, we refer the reader to~\cite{Bresson2017}.

Neural networks such as PoseNet~\cite{Kendall2015ICCV}, VLocNet++~\cite{Radwan2018VLocNet}, or the approaches introduced in~\cite{Walch2017ImageBasedLU},~\cite{Melekhov2017ImageBasedLU},~\cite{LMKK2017ICCVW},~\cite{Brachmann2018CVPR}, or~\cite{Sarlin2018} are using image data to estimate the 3D pose of a camera in an End2End fashion. Scene semantics can be derived together with the estimated pose~\cite{Radwan2018VLocNet}.

LiDAR intensity maps are also suited for learning a real-time, calibration-agnostic localization for autonomous cars~\cite{Barsan2018}. The method uses a deep neural network to build a learned representation of the driving scene from LiDAR sweeps and intensity maps. The localization of the vehicle is obtained through convolutional matching. In~\cite{garcia2009laser}, laser scans and a deep neural network are used to learn descriptors for localization in urban and natural environments.

In order to safely navigate the driving scene, an autonomous car should be able to estimate the motion of the surrounding environment, also known as \textit{scene flow}. Previous LiDAR based scene flow estimation techniques mainly relied on manually designed features. In recent articles, we have noticed a tendency to replace these classical methods with deep learning architectures able to automatically learn the scene flow. In~\cite{pmlr-v87-ushani18a}, an encoding deep network is trained on occupancy grids with the purpose of finding matching or non-matching locations between successive timesteps.

Although much progress has been reported in the area of deep learning based localization, VO techniques are still dominated by classical keypoints matching algorithms, combined with acceleration data provided by inertial sensors. This is mainly due to the fact that keypoints detectors are computational efficient and can be easily deployed on embedded devices.

\subsection{Perception using Occupancy Maps}

\begin{figure*}
	\centering
	\begin{center}
		\includegraphics[scale=1.07]{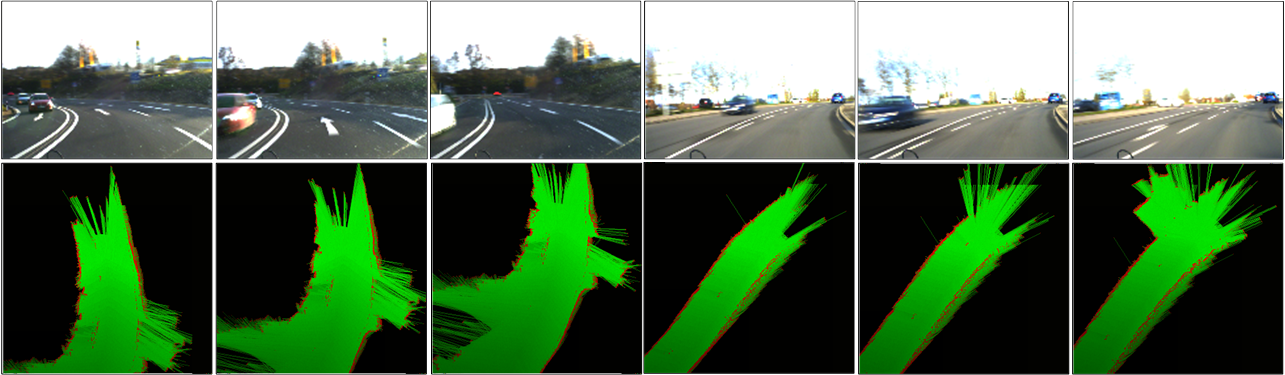}
		\caption{\textbf{Examples of Occupancy Grids (OG).} The images show a snapshot of the driving environment together with its respective occupancy grid~\cite{Marina_IRC2019}.}
        \label{fig:ROB-19-0022_occupancy_grids_fig6}
	\end{center}
\end{figure*}

An occupancy map, also known as Occupancy Grid (OG), is a representation of the environment which divides the driving space into a set of cells and calculates the occupancy probability for each cell. Popular in robotics \cite{garcia2009laser},~\cite{ProbabilThrun}, the OG representation became a suitable solution for self-driving vehicles. A couple of OG data samples are shown in Fig.~\ref{fig:ROB-19-0022_occupancy_grids_fig6}.

Deep learning is used in the context of occupancy maps either for dynamic objects detection and tracking~\cite{OndruskaDWP16}, probabilistic estimation of the occupancy map surrounding the vehicle~\cite{HoermannBD17},\cite{RamosGPFR16}, or for deriving the driving scene context~\cite{Seeger},~\cite{Marina_IRC2019}. In the latter case, the OG is constructed by accumulating data over time, while a deep neural net is used to label the environment into driving context classes, such as highway driving, parking area, or inner-city driving.

Occupancy maps represent an in-vehicle virtual environment, integrating perceptual information in a form better suited for path planning and motion control. Deep learning plays an important role in the estimation of OG, since the information used to populate the grid cells is inferred from processing image and LiDAR data using scene perception methods, as the ones described in this chapter of the survey.

\section{Deep Learning for Path Planning and Behavior Arbitration}
\label{sec:path_planning}

The ability of an autonomous car to find a route between two points, that is, a start position and a desired location, represents \textit{path planning}. According to the path planning process, a self-driving car should consider all possible obstacles that are present in the surrounding environment and calculate a trajectory along a collision-free route. As stated in \cite{shalevshwartz2016safe}, autonomous driving is a multi-agent setting where the host vehicle must apply sophisticated negotiation skills with other road users when overtaking, giving way, merging, taking left and right turns, all while navigating unstructured urban roadways. The literature findings point to a non trivial policy that should handle safety in driving. Considering a reward function $R(\bar{s})=-r$ for an accident event that should be avoided and $R(\bar{s})\in [-1,1]$ for the rest of the trajectories, the goal is to learn to perform difficult maneuvers smoothly and safe.

This emerging topic of optimal path planning for autonomous cars should operate at high computation speeds, in order to obtain short reaction times, while satisfying specific optimization criteria. The survey in \cite{Pendleton2017} provides a general overview of path planning in the automotive context. It addresses the taxonomy aspects of path planning, namely the mission planner, behavior planner and motion planner. However,~\cite{Pendleton2017} does not include a review on deep learning technologies, although the state of the art literature has revealed an increased interest in using deep learning technologies for path planning and behavior arbitration. Following, we discuss two of the most representative deep learning paradigms for path planning, namely Imitation Learning (IL)~\cite{Rehder2017},~\cite{Liting2017},~\cite{Grigorescu2019RAL} and Deep Reinforcement Learning (DRL) based planning~\cite{Yu2018}~\cite{Paxton2017}.

The goal in \textit{Imitation Learning}~\cite{Rehder2017},~\cite{Liting2017},~\cite{Grigorescu2019RAL} is to learn the behavior of a human driver from recorded driving experiences~\cite{Schwarting2018}. The strategy implies a vehicle teaching process from human demonstration. Thus, the authors employ CNNs to learn planning from imitation. For example, NeuroTrajectory~\cite{Grigorescu2019RAL} is a perception-planning deep neural network that learns the desired state trajectory of the ego-vehicle over a finite prediction horizon. Imitation learning can also be framed as an Inverse Reinforcement Learning (IRL) problem, where the goal is to learn the reward function from a human driver~\cite{Gu2016},~\cite{Wulfmeier2016}. Such methods use real drivers behaviors to learn reward-functions and to generate human-like driving trajectories.

\textit{DRL for path planning} deals mainly with learning driving trajectories in a simulator~\cite{shalevshwartz2016safe},~\cite{Aleksandr2018},~\cite{Yu2018}~\cite{Paxton2017}. The real environmental model is abstracted and transformed into a virtual environment, based on a transfer model.
In~\cite{shalevshwartz2016safe}, it is stated that the objective function cannot ensure functional safety without causing a serious variance problem. The proposed solution for this issue is to construct a policy function composed of learnable and non-learnable parts. The learnable policy tries to maximize a reward function (which includes comfort, safety, overtake opportunity, etc.). At the same time, the non-learnable policy follows the hard constraints of functional safety, while maintaining an acceptable level of comfort.

Both IL and DRL for path planning have advantages and disadvantages. IL has the advantage that it can be trained with data collected from the real-world. Nevertheless, this data is scarce on corner cases (e.g. driving off-lanes, vehicle crashes, etc.), making the trained network's response uncertain when confronted with unseen data. On the other hand, although DRL systems are able to explore different driving situations within a simulated world, these models tend to have a biased behavior when ported to the real-world.

\section{Motion Controllers for AI-based Self-Driving Cars}
\label{sec:controllers}

The motion controller is responsible for computing the longitudinal and lateral steering commands of the vehicle. Learning algorithms are used either as part of \textit{Learning Controllers}, within the motion control module from Fig.~\ref{fig:ROB-19-0022_standard_had_system_fig1}(a), or as complete \textit{End2End Control Systems} which directly map sensory data to steering commands, as shown in Fig.~\ref{fig:ROB-19-0022_standard_had_system_fig1}(b).

\subsection{Learning Controllers}
\label{sec:learning_controllers}

Traditional controllers make use of an \textit{a priori} model composed of fixed parameters. When robots or other autonomous systems are used in complex environments, such as driving, traditional controllers cannot foresee every possible situation that the system has to cope with. Unlike controllers with fixed parameters, learning controllers make use of training information to learn their models over time. With every gathered batch of training data, the approximation of the true system model becomes more accurate, thus enabling model flexibility, consistent uncertainty estimates and anticipation of repeatable effects and disturbances that cannot be modeled prior to deployment~\cite{ostafew-jfr15}. Consider the following nonlinear, state-space system:

\begin{equation}
	\vec{z}^{<t+1>} = \vec{f}_{true} (\vec{z}^{<t>}, \vec{u}^{<t>}),
\end{equation}

\noindent with observable state $\vec{z}^{<t>} \in \mathbb{R}^n$ and control input $\vec{u}^{<t>} \in \mathbb{R}^m$, at discrete time $t$. The true system $\vec{f}_{true}$ is not known exactly and is approximated by the sum of an \textit{a-priori} model and a learned dynamics model:

\begin{equation}
	\vec{z}^{<t+1>} = \underset{\text{\textit{a-priori} model}}{\vec{f} (\vec{z}^{<t>}, \vec{u}^{<t>})} + \underset{\text{learned model}}{\vec{h} (\vec{z}^{<t>})}.
\end{equation}

In previous works, learning controllers have been introduced based on simple function approximators, such as Gaussian Process (GP) modeling~\cite{Nguyen2008}, \cite{Meier_IROS_2014}, \cite{ostafew-jfr15}, \cite{ostafew-ijrr16}, or Support Vector Regression~\cite{Sigaud_2011}.

Learning techniques are commonly used to learn a dynamics model which in turn improves an a priori system model in \textit{Iterative Learning Control} (ILC)~\cite{Ostafew2013},~\cite{Panomruttanarug2017},~\cite{Kapania2015},~\cite{YangZhao2017} and \textit{Model Predictive Control} (MPC)~\cite{Lefevre2016_1}~\cite{Lefevre2016_2},~\cite{ostafew-jfr15},~\cite{ostafew-ijrr16},~\cite{Drews2017_1},~\cite{Drews2017_2},~\cite{Rosolia2017},~\cite{Pan2017},~\cite{Pan2018}.

\textit{Iterative Learning Control} (ILC) is a method for controlling systems which work in a repetitive mode, such as path tracking in self-driving cars. It has been successfully applied to navigation in off-road terrain~\cite{Ostafew2013}, autonomous car parking~\cite{Panomruttanarug2017} and modeling of steering dynamics in an autonomous race car~\cite{Kapania2015}. Multiple benefits are highlighted, such as the usage of a simple and computationally light feedback controller, as well as a decreased controller design effort (achieved by predicting path disturbances and platform dynamics).

\textit{Model Predictive Control} (MPC)~\cite{rawlings2009model} is a control strategy that computes control actions by solving an optimization problem. It received lots of attention in the last two decades due to its ability to handle complex nonlinear systems with state and input constraints. The central idea behind MPC is to calculate control actions at each sampling time by minimizing a cost function over a short time horizon, while considering observations, input-output constraints and the system's dynamics given by a process model. A general review of MPC techniques for autonomous robots is given in~\cite{Kamel2018}.

Learning has been used in conjunction with MPC to learn driving models~\cite{Lefevre2016_1},~\cite{Lefevre2016_2}, driving dynamics for race cars operating at their handling limits~\cite{Drews2017_1},~\cite{Drews2017_2},~\cite{Rosolia2017}, as well as to improve path tracking accuracy~\cite{Brunner2017},~\cite{ostafew-jfr15},~\cite{ostafew-ijrr16}. These methods use learning mechanisms to identify nonlinear dynamics that are used in the MPC's trajectory cost function optimization. This enables one to better predict disturbances and the behavior of the vehicle, leading to optimal comfort and safety constraints applied to the control inputs. Training data is usually in the form of past vehicle states and observations. For example, CNNs can be used to compute a dense occupancy grid map in a local robot-centric coordinate system. The grid map is further passed to the MPC's cost function for optimizing the trajectory of the vehicle over a finite prediction horizon.

A major advantage of learning controllers is that they optimally combine traditional model-based control theory with learning algorithms. This makes it possible to still use established methodologies for controller design and stability analysis, together with a robust learning component applied at system identification and prediction levels.

\subsection{End2End Learning Control}
\label{sec:end2end_learning}

In the context of autonomous driving, End2End Learning Control is defined as a direct mapping from sensory data to control commands. The inputs are usually from a high-dimensional features space (e.g. images or point clouds). As illustrated in Fig~\ref{fig:ROB-19-0022_standard_had_system_fig1}(b), this is opposed to traditional processing pipelines, where at first objects are detected in the input image, after which a path is planned and finally the computed control values are executed. A summary of some of the most popular End2End learning systems is given in Table~\ref{end2end_table}.

\begin{table*}[!h]\centering
\resizebox{\textwidth}{!}{\begin{tabular}{ccccl@{}}
\toprule
\rowcolor[HTML]{E7E7E7} 
\textbf{Name} & \textbf{Problem Space} & \textbf{\begin{tabular}[c]{@{}c@{}}Neural network \\ architecture\end{tabular}} & \textbf{\begin{tabular}[c]{@{}c@{}}Sensor\\ input\end{tabular}} & \multicolumn{1}{c}{\cellcolor[HTML]{E7E7E7}\textbf{Description}} \\ \midrule
\begin{tabular}[c]{@{}c@{}}ALVINN\\~\cite{pomerleau1989alvinn}\end{tabular} & Road following & \begin{tabular}[c]{@{}c@{}}3-layer\\ back-prop.\\ network\end{tabular} & \begin{tabular}[c]{@{}c@{}}Camera, laser\\ range finder\end{tabular} & \begin{tabular}[c]{@{}l@{}}ALVINN stands for Autonomous Land Vehicle In a Neural \\ Network). Training has been conducted using simulated \\ road images. Successful tests on the Carnegie Mellon \\ autonomous navigation test vehicle indicate that the\\ network can effectively follow real roads.\end{tabular} \\ \midrule
\begin{tabular}[c]{@{}c@{}}DAVE\\~\cite{muller2006off}\end{tabular} & DARPA challenge & \begin{tabular}[c]{@{}c@{}}6-layer \\ CNN\end{tabular} & \begin{tabular}[c]{@{}c@{}}Raw camera\\ images\end{tabular} & \begin{tabular}[c]{@{}l@{}}A vision-based obstacle avoidance system for off-road \\mobile robots. The robot is a 50cm off-road truck, with two \\front color cameras. A remote computer processes the video \\and controls the robot via radio.\end{tabular} \\ \midrule
\begin{tabular}[c]{@{}c@{}}NVIDIA PilotNet\\~\cite{bojarski2017explaining}\end{tabular} & \begin{tabular}[c]{@{}c@{}}Autonomous \\driving in real \\traffic situations\end{tabular} & CNN & \begin{tabular}[c]{@{}c@{}}Raw camera\\ images\end{tabular} & \begin{tabular}[c]{@{}l@{}}The system automatically learns internal representations of \\the necessary processing steps such as detecting useful road \\ features with human steering angle as the training signal.\end{tabular} \\ \midrule
\begin{tabular}[c]{@{}c@{}}Novel FCN-LSTM\\~\cite{xu2017end}\end{tabular} & \begin{tabular}[c]{@{}c@{}}Ego-motion \\prediction\end{tabular} & FCN-LSTM & \begin{tabular}[c]{@{}c@{}}Large scale\\ video data\end{tabular} & \begin{tabular}[c]{@{}l@{}}A generic vehicle motion model from large scale crowd-\\sourced video data is obtained, while developing an end-to\\-end trainable architecture (FCN-LSTM) for predicting a \\distribution of future vehicle ego-motion data.\end{tabular} \\ \midrule
\begin{tabular}[c]{@{}c@{}}Novel C-LSTM\\~\cite{eraqi2017end}\end{tabular} & Steering angle control & C-LSTM & \begin{tabular}[c]{@{}c@{}}Camera frames,\\ steering wheel\\ angle\end{tabular} & \begin{tabular}[c]{@{}l@{}}C-LSTM is end-to-end trainable, learning both visual and \\ dynamic temporal dependencies of driving. Additionally, the \\ steering angle regression problem is considered classification \\while imposing a spatial relationship between the output\\ layer neurons.\end{tabular} \\ \midrule
\begin{tabular}[c]{@{}c@{}}Drive360\\~\cite{hecker2018end}\end{tabular} & \begin{tabular}[c]{@{}c@{}}Steering angle and \\ velocity control\end{tabular} & \begin{tabular}[c]{@{}c@{}}CNN + Fully\\ Connected + \\ LSTM\end{tabular} & \begin{tabular}[c]{@{}c@{}}Surround-view\\ cameras, CAN\\ bus reader\end{tabular} & \begin{tabular}[c]{@{}l@{}}The sensor setup provides data for a 360-degree view of \\ the area surrounding the vehicle. A new driving dataset \\ is collected, covering diverse scenarios. A novel driving \\ model is developed by integrating the surround-view \\ cameras with the route planner.\end{tabular} \\ \midrule
\begin{tabular}[c]{@{}c@{}}DNN policy\\~\cite{Rausch2017}\end{tabular} & Steering angle control & CNN + FC & Camera images & \begin{tabular}[c]{@{}l@{}}The trained neural net directly maps pixel data from a \\ front-facing camera to steering commands and does not \\ require any other sensors. We compare the controller\\performance with the steering behavior of a human driver.\end{tabular} \\ \midrule
\begin{tabular}[c]{@{}c@{}}DeepPicar\\~\cite{Bechtel2018}\end{tabular} & Steering angle control & CNN & Camera images & \begin{tabular}[c]{@{}l@{}}DeepPicar is a small scale replica of a real self-driving car \\ called DAVE-2 by NVIDIA. It uses the same network \\ architecture and can drive itself in real-time using a web \\ camera and a Raspberry Pi 3.\end{tabular} \\ \midrule
\begin{tabular}[c]{@{}c@{}}TORCS DRL\\~\cite{Sallab2017}\end{tabular} & \begin{tabular}[c]{@{}c@{}}Lane keeping and \\ obstacle avoidance\end{tabular} & \begin{tabular}[c]{@{}c@{}}DQN + RNN \\ + CNN\end{tabular} & \begin{tabular}[c]{@{}c@{}}TORCS \\ simulator\\ images\end{tabular} & \begin{tabular}[c]{@{}l@{}}It incorporates Recurrent Neural Networks for information\\ integration, enabling the car to handle partially observable \\ scenarios. It also reduces the computational complexity for\\ deployment on embedded hardware.\end{tabular} \\ \midrule
\begin{tabular}[c]{@{}c@{}}TORCS E2E\\~\cite{Yang2017}\end{tabular} & \begin{tabular}[c]{@{}c@{}}Steering angle control\\ in a simulated\\ env. (TORCS)\end{tabular} & CNN & \begin{tabular}[c]{@{}c@{}}TORCS\\ simulator\\ images\end{tabular} & \begin{tabular}[c]{@{}l@{}}The image features are split into three categories (sky-related, \\ roadside-related, and roadrelated features). Two experimental\\ frameworks are used to investigate the importance of each \\ single feature for training a CNN controller.\end{tabular} \\ \midrule
\begin{tabular}[c]{@{}c@{}}Agile Autonomous Driving\\~\cite{Pan2018}\end{tabular} & \begin{tabular}[c]{@{}c@{}}Steering angle and\\velocity control\\for aggressive driving\end{tabular} & CNN & \begin{tabular}[c]{@{}c@{}}Raw camera\\images\\ \end{tabular} & \begin{tabular}[c]{@{}l@{}} A CNN, refereed to as the learner, is trained with optimal \\ trajectory examples provided at training time by an MPC controller. \\ The MPC acts as an expert, encoding the scene dynamics \\ into the layers of the neural network. \end{tabular} \\ \midrule
\begin{tabular}[c]{@{}c@{}}WRC6 AD\\~\cite{Jaritz2018}\end{tabular} & \begin{tabular}[c]{@{}c@{}}Driving in a \\ racing game\end{tabular} & \begin{tabular}[c]{@{}c@{}}CNN + LSTM\\ Encoder\end{tabular} & \begin{tabular}[c]{@{}c@{}}WRC6 \\ Racing\\ Game\end{tabular} & \begin{tabular}[c]{@{}l@{}}An Asynchronous ActorCritic (A3C) framework is used to \\ learn the car control in a physically and graphically realistic \\ rally game, with the agents evolving simultaneously on \\ different tracks.\end{tabular} \\ \bottomrule
\end{tabular}}
\caption{Summary of End2End learning methods.}
\label{end2end_table}
\end{table*}

End2End learning can also be formulated as a back-propagation algorithm scaled up to complex models. The paradigm was first introduced in the 1990s, when the Autonomous Land Vehicle in a Neural Network (ALVINN) system was built~\cite{pomerleau1989alvinn}. ALVINN was designed to follow a pre-defined road, steering according to the observed road's curvature. The next milestone in End2End driving is considered to be in the mid 2000s, when DAVE (Darpa Autonomous VEhicle) managed to drive through an obstacle-filled road, after it has been trained on hours of human driving acquired in similar, but not identical, driving scenarios~\cite{muller2006off}. Over the last couple of years, the technological advances in computing hardware have facilitated the usage of End2End learning models. The back-propagation algorithm for gradient estimation in deep networks is now efficiently implemented on parallel Graphic Processing Units (GPUs). This kind of processing allows the training of large and complex network architectures, which in turn require huge amounts of training samples (see Section~\ref{sec:data_sources}). 

End2End control papers mainly employ either deep neural networks trained offline on real-world and/or synthetic data~\cite{Bojarski2016},~\cite{xu2017end},~\cite{eraqi2017end},~\cite{hecker2018end},~\cite{Fridman2017},~\cite{Rausch2017},~\cite{Bechtel2018},~\cite{Chen2015},~\cite{Yang2017}, or Deep Reinforcement Learning (DRL) systems trained and evaluated in simulation~\cite{Sallab2017}~\cite{Perot2017},~\cite{Jaritz2018}. Methods for porting simulation trained DRL models to real-world driving have also been reported~\cite{Wayve2018}, as well as DRL systems trained directly on real-world image data~\cite{Pan2017},~\cite{Pan2018}.

End2End methods have been popularized in the last couple of years by NVIDIA\textsuperscript{\textregistered}, as part of the \textit{PilotNet} architecture. The approach is to train a CNN which maps raw pixels from a single front-facing camera directly to steering commands~\cite{Bojarski2016}. The training data is composed of images and steering commands collected in driving scenarios performed in a diverse set of lighting and weather conditions, as well as on different road types. Prior to training, the data is enriched using augmentation, adding artificial shifts and rotations to the original data.

PilotNet has $250.000$ parameters and approx. $27mil.$ connections. The evaluation is performed in two stages: first in simulation and secondly in a test car. An \textit{autonomy} performance metric represents the percentage of time when the neural network drives the car:

\begin{equation}
	autonomy = (1 - \frac{(no.\;of\;interventions) * 6\;sec}{elapsed\; time\;[sec]})*100.
\end{equation}

An intervention is considered to take place when the simulated vehicle departs from the center line by more than one meter, assuming that $6$ seconds is the time needed by a human to retake control of the vehicle and bring it back to the desired state. An autonomy of $98\%$ was reached on a $20km$ drive from Holmdel to Atlantic Highlands in NJ, USA. Through training, PilotNet learns how the steering commands are computed by a human driver~\cite{bojarski2017explaining}. The focus is on determining which elements in the input traffic image have the most influence on the network's steering decision. A method for finding the salient object regions in the input image is described, while reaching the conclusion that the low-level features learned by PilotNet are similar to the ones that are relevant to a human driver.

End2End architectures similar to PilotNet, which map visual data to steering commands, have been reported in~\cite{Rausch2017},~\cite{Bechtel2018},~\cite{Chen2015}. In~\cite{xu2017end}, autonomous driving is formulated as a future ego-motion prediction problem. The introduced FCN-LSTM (Fully Convolutional Network - Long-Short Term Memory) method is designed to jointly train pixel-level supervised tasks using a fully convolutional encoder, together with motion prediction through a temporal encoder. The combination between visual temporal dependencies of the input data has also been considered in~\cite{eraqi2017end}, where the C-LSTM (Convolutional Long Short Term Memory) network has been proposed for steering control. In~\cite{hecker2018end}, surround-view cameras were used for End2End learning. The claim is that human drivers also use rear and side-view mirrors for driving, thus all the information from around the vehicle needs to be gathered and integrated into the network model in order to output a suitable control command.

To carry out an evaluation of the Tesla\textsuperscript{\textregistered} Autopilot system,~\cite{Fridman2017} proposed an End2End Convolutional Neural Network framework. It is designed to determine differences between Autopilot and its own output, taking into consideration edge cases. The network was trained using real data, collected from over $420$ hours of real road driving. The comparison between Tesla\textsuperscript{\textregistered}'s Autopilot and the proposed framework was done in real-time on a Tesla\textsuperscript{\textregistered} car. The evaluation revealed an accuracy of $90.4\%$ in detecting differences between both systems and the control transfer of the car to a human driver. 

Another approach to design End2End driving systems is DRL. This is mainly performed in simulation, where an autonomous agent can safely explore different driving strategies. In~\cite{Sallab2017}, a DRL End2End system is used to compute steering command in the TORCS game simulation engine. Considering a more complex virtual environment,~\cite{Perot2017} proposed an asynchronous advantage Actor-Critic (A3C) method for training a CNN on images and vehicle velocity information. The same idea has been enhanced in~\cite{Jaritz2018}, having a faster convergence and permissiveness for more generalization. Both articles rely on the following procedure: receiving the current state of the game, deciding on the next control commands and then getting a reward on the next iteration. The experimental setup benefited from a realistic car game, namely World Rally Championship 6, and also from other simulated environments, like TORCS.

The next trend in DRL based control seems to be the inclusion of classical model-based control techniques, as the ones detailed in Section~\ref{sec:learning_controllers}. The classical controller provides a stable and deterministic model on top of which the policy of the neural network is estimated. In this way, the hard constraints of the modeled system are transfered into the neural network policy~\cite{Zhang_2016}. A DRL policy trained on real-world image data has been proposed in~\cite{Pan2017} and~\cite{Pan2018} for the task of aggressive driving. In this case, a CNN, refereed to as the learner, is trained with optimal trajectory examples provided at training time by a model predictive controller.

\section{Safety of Deep Learning in Autonomous Driving}
\label{sec:safety}

Safety implies the absence of the conditions that cause a system to be dangerous~\cite{ferrel2010}. Demonstrating the safety of a system which is running deep learning techniques depends heavily on the type of technique and the application context. Thus, reasoning about the safety of deep learning techniques requires:

\begin{itemize}
	\item understanding the impact of the possible failures;
	\item understanding the context within the wider system;
	\item defining the assumption regarding the system context and the environment in which it will likely be used;
	\item defining what a safe behavior means, including non-functional constraints.
\end{itemize}

In~\cite{Burton2017}, an example is mapped on the above requirements with respect to a deep learning component. The problem space for the component is pedestrian detection with convolutional neural networks. The top level task of the system is to locate an object of class person from a distance of 100 meters, with a lateral accuracy of +/- 20 cm, a false negative rate of 1\% and false positive rate of 5\%. The assumptions is that the braking distance and speed are sufficient to react when detecting persons which are 100 meters ahead of the planned trajectory of the vehicle. Alternative sensing methods can be used in order to reduce the overall false negative and false positive rates of the system to an acceptable level. The context information is that the distance and the accuracy shall be mapped to the dimensions of the image frames presented to the CNN.  

There is no commonly agreed definition for the term \textit{safety} in the context of machine learning or deep learning. In \cite{Varshney2016}, Varshney defines safety in terms of risk, epistemic uncertainty and the harm incurred by unwanted outcomes. He then analyses the choice of cost function and the appropriateness of minimizing the empirical average training cost.

\cite{Amodei2016ConcretePI} takes into consideration the problem of \textit{accidents} in machine learning systems. Such accidents are defined as unintended and harmful behaviors that may emerge from a poor AI system design. The authors present a list of five practical research problems related to accident risk, categorized according to whether the problem originates from having the wrong objective function (\textit{avoiding side effects} and \textit{avoiding reward hacking}), an objective function that is too expensive to evaluate frequently (\textit{scalable supervision}), or undesirable behavior during the learning process (\textit{safe exploration} and \textit{distributional shift}).

Enlarging the scope of safety, \cite{Möller2012} propose a decision-theoretic definition of safety that applies to a broad set of domains and systems. They define safety to be the reduction or minimization of risk and epistemic uncertainty associated with unwanted outcomes that are severe enough to be seen as harmful. The key points in this definition are: \textit{i}) the cost of unwanted outcomes has to be sufficiently high in some human sense for events to be harmful, and \textit{ii}) safety involves reducing both the probability of expected harms, as well as the possibility of unexpected harms.

Regardless of the above empirical definitions and possible interpretations of safety, the use of deep learning components in safety critical systems is still an open question. The ISO 26262 standard for functional safety of road vehicles provides a comprehensive set of requirements for assuring safety, but does not address the unique characteristics of deep learning-based software. 

\cite{Salay2017} addresses this gap by analyzing the places where machine learning can impact the standard and provides recommendations on how to accommodate this impact. These recommendations are focused towards the direction of identifying the hazards, implementing tools and mechanism for fault and failure situations, but also ensuring complete training datasets and designing a multi-level architecture. The usage of specific techniques for various stages within the software development life-cycle is desired. 

The standard ISO 26262 recommends the use of a Hazard Analysis and Risk Assessment (HARA) method to identify hazardous events in the system and to specify safety goals that mitigate the hazards. The standard has 10 parts. Our focus is on Part 6: \textit{product development at the software level}, the standard following the well-known V model for engineering. Automotive Safety Integrity Level (ASIL) refers to a risk classification scheme defined in ISO 26262 for an item (e.g. subsystem) in an automotive system. 

ASIL represents the degree of rigor required (e.g., testing techniques, types of documentation required, etc.) to reduce risk, where ASIL D represents the highest and ASIL A the lowest risk. If an element is assigned to QM (Quality Management), it does not require safety management. The ASIL assessed for a given hazard is at first assigned to the safety goal set to address the hazard and is then inherited by the safety requirements derived from that goal~\cite{Salay2017}.

According to ISO26226, a \textit{hazard} is defined as "potential source of harm caused by a malfunctioning behavior, where harm is a physical injury or damage to the health of a person"~\cite{Spanfeiner2012}. Nevertheless, a deep learning component can create new types of hazards. An example of such a hazard is usually happening because humans think that the automated driver assistance (often developed using learning techniques) is more reliable than it actually is~\cite{Parasuraman1997}.

Due to its complexity, a deep learning component can fail in unique ways. For example, in Deep Reinforcement Learning systems, faults in the reward function can negatively affect the trained model~\cite{Amodei2016ConcretePI}. In such a case, the automated vehicle figures out that it can avoid getting penalized for driving too close to other vehicles by exploiting certain sensor vulnerabilities so that it \textit{can't see} how close it is getting. Although hazards such as these may be unique to deep reinforcement learning components, they can be traced to faults, thus fitting within the existing guidelines of ISO 26262.

A key requirement for analyzing the safety of deep learning components is to examine whether immediate human costs of outcomes exceed some harm severity thresholds. Undesired outcomes are truly harmful in a human sense and their effect is felt in near real-time. These outcomes can be classified as safety issues. The cost of deep learning decisions is related to optimization formulations which explicitly include a loss function $L$. The loss function $L : X \times Y \times Y → \rightarrow R$ is defined as the measure of the error incurred by predicting the label of an observation $x$ as $f(x)$, instead of $y$. Statistical learning calls the risk of $f$ as the expected value of the loss of $f$ under $P$:

\begin{equation}
	R(f) = \int_{} L(x,f(x), y)dP(x,y),
\end{equation}
	
\noindent where, $X \times Y$ is a random example space of observations $x$ and labels $y$, distributed according to a probability distribution $P(X,Y)$. The statistical learning problem consists of finding the function $f$ that optimizes (i.e. minimizes) the risk $R$~\cite{Faria2018}. For an algorithm's hypothesis $h$ and loss function $L$, the expected loss on the training set is called the \textit{empirical risk} of $h$:

\begin{equation}
	\vec{R}_{emp} (h) = \frac{1}{m} \sum\limits_{i=1}^{m} L(x^{(i)}, h(x)^{(i)}, y^{(i)}).
\end{equation}

\noindent A machine learning algorithm then optimizes the empirical risk on the expectation that the risk decreases significantly. However, this standard formulation does not consider the issues related to the uncertainty that is relevant for safety. The distribution of the training samples ${(x_1, y_1),...,(x_m, y_m)}$ is drawn from the true underlying probability distribution of $(X, Y)$, which may not always be the case. Usually the probability distribution is unknown, precluding the use of domain adaptation techniques~\cite{Daume2006}~\cite{Caruana2015}. This is one of the epistemic uncertainty that is relevant for safety because training on a dataset of different distribution can cause much harm through bias.

In reality, a machine learning system only encounters a finite number of test samples and an actual operational risk is an empirical quantity on the test set. The operational risk may be much larger than the actual risk for small cardinality test sets, even if $h$ is risk-optimal. This uncertainty caused by the instantiation of the test set can have large safety implications on individual test samples~\cite{VarshneyMLSafety2016}. 

Faults and failures of a programmed component (e.g. one using a formal algorithm to solve a problem) are totally different from the ones of a deep learning component. Specific faults of a deep learning component can be caused by unreliable or noisy sensor signals (video signal due to bad weather, radar signal due to absorbing construction materials, GPS data, etc.), neural network topology, learning algorithm, training set or unexpected changes in the environment (e.g. unknown driving scenes or accidents on the road). We must mention the first autonomous driving accident, produced by a Tesla\textsuperscript{\textregistered} car, where, due to object misclassification errors, the AutoPilot function collided the vehicle into a truck~\cite{Levin2018}. Despite the 130 million miles of testing and evaluation, the accident was caused under extremely rare circumstances, also known as Black Swans, given the height of the truck, its white color under bright sky, combined with the positioning of the vehicle across the road.

Self-driving vehicles must have fail-safe mechanisms, usually encountered under the name of \textit{Safety Monitors}. These must stop the autonomous control software once a failure is detected~\cite{Koopman2017}. Specific fault types and failures have been cataloged for neural networks in~\cite{Kurd2007},~\cite{Harris2016} and~\cite{McPherson2018}. This led to the development of specific and focused tools and techniques to help finding faults. \cite{Chakarov2018} describes a technique for debugging misclassifications due to bad training data, while an approach for troubleshooting faults due to complex interactions between linked machine learning components is proposed in~\cite{Nushi2017OnHI}. In~\cite{Takanami2000}, a white box technique is used to inject faults onto a neural network by breaking the links or randomly changing the weights.

The training set plays a key role in the safety of the deep learning component. ISO 26262 standard states that the component behavior shall be fully specified and each refinement shall be verified with respect to its specification. This assumption is violated in the case of a deep learning system, where a training set is used instead of a specification. It is not clear how to ensure that the corresponding hazards are always mitigated. The training process is not a verification process since the trained model will be “correct by construction” with respect to the training set, up to the limits of the model and the learning algorithm~\cite{Salay2017}. Effects of this considerations are visible in the commercial autonomous vehicle market, where Black Swan events caused by data not present in the training set may lead to fatalities~\cite{McPherson2018}. 

Detailed requirements shall be formulated and traced to hazards. Such a requirement can specify how the training, validation and testing sets are obtained. Subsequently, the data gathered can be verified with respect to this specification. Furthermore, some specifications, for example the fact that a vehicle cannot be wider than 3 meters, can be used to reject false positive detections. Such properties are used even directly during the training process to improve the accuracy of the model~\cite{Katz2017}. 

Machine learning and deep learning techniques are starting to become effective and reliable even for safety critical systems, even if the complete safety assurance for this type of systems is still an open question. Current standards and regulation from the automotive industry cannot be fully mapped to such systems, requiring the development of new safety standards targeted for deep learning.

\section{Data Sources for Training Autonomous Driving Systems}
\label{sec:data_sources}

\begin{figure}
	\centering
	\begin{center}
		\includegraphics[scale=0.235]{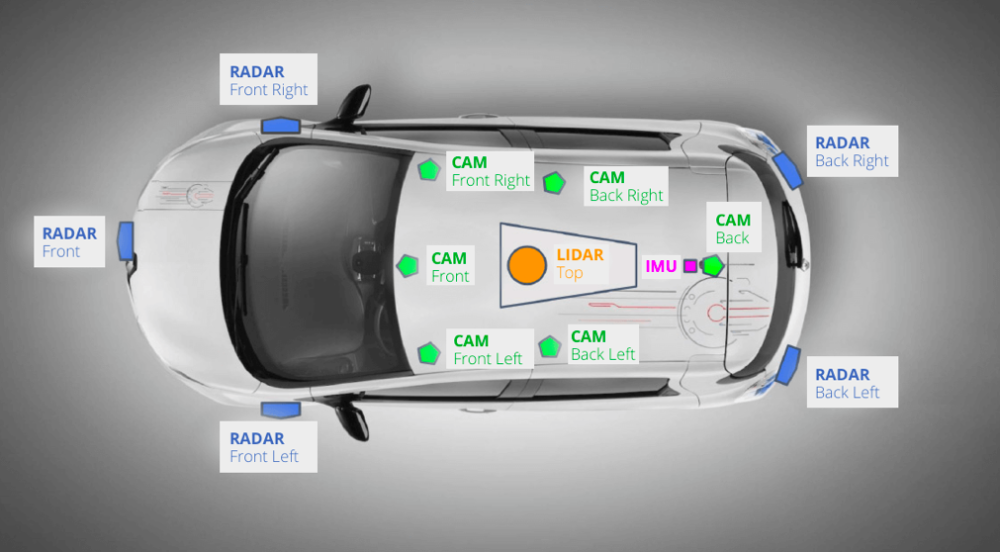}
		\caption{\textbf{Sensor suite of the nuTonomy\textsuperscript{\textregistered} self-driving car \cite{nuscenes2019}}.}
        \label{fig:ROB-19-0022_Sensors_setup_nuScenes_fig7}
	\end{center}
\end{figure}

Undeniably, the usage of real world data is a key requirement for training and testing an autonomous driving component. The high amount of data needed in the development stage of such components made data collection on public roads a valuable activity. In order to obtain a comprehensive description of the driving scene, the vehicle used for data collection is equipped with a variety of sensors such as radar, LIDAR, GPS, cameras, Inertial Measurement Units (IMU) and ultrasonic sensors. The sensors setup differs from vehicle to vehicle, depending on how the data is planned to be used. A common sensor setup for an autonomous vehicle is presented in Fig.~\ref{fig:ROB-19-0022_Sensors_setup_nuScenes_fig7}.

In the last years, mainly due to the large and increasing research interest in autonomous vehicles, many driving datasets were made public and documented. They vary in size, sensor setup and data format. The researchers need only to identify the proper dataset which best fits their problem space. \cite{Janai2017} published a survey on a broad spectrum of datasets. These datasets address the computer vision field in general, but there are few of them which fit to the autonomous driving topic. 

A most comprehensive survey on publicly available datasets for self-driving vehicles algorithms can be found in~\cite{Yin2017}. The paper presents 27 available datasets containing data recorded on public roads. The datasets are compared from different perspectives, such that the reader can select the one best suited for his task. 

Despite our extensive search, we are yet to find a master dataset that combines at least parts of the ones available. The reason may be that there are no standard requirements for the data format and sensor setup. Each dataset heavily depends on the objective of the algorithm for which the data was collected. Recently, the companies Scale\textsuperscript{\textregistered} and nuTonomy\textsuperscript{\textregistered} started to create one of the largest and most detailed self-driving dataset on the market to date\footnote{\url{https://venturebeat.com/2018/09/14/scale-and-nutonomy-release-nuscenes-a-self-driving-dataset-with-over-1-4-million-images/}}. This includes Berkeley DeepDrive~\cite{DeepDrive_Yu_2018}, a dataset developed by researchers at Berkeley University. More relevant datasets from the literature are pending for merging\footnote{\url{https://scale.com/open-datasets}}.

In~\cite{Fridman2017}, the authors present a study that seeks to collect and analyze large scale naturalistic data of semi-autonomous driving in order to better characterize the state of the art of the current technology. The study involved $99$ participants, $29$ vehicles, $405,807$ miles and approximatively $5.5$ billion video frames. Unfortunately, the data collected in this study is not available for the public.

In the remaining of this section we will provide and highlight the distinctive characteristics of the most relevant datasets that are publicly available.

\begin{table*}[!h] \centering
\resizebox{\textwidth}{!}{\begin{tabular}{ccccccc}
\hline
\rowcolor[HTML]{E7E7E7} 
\textbf{Dataset} & \textbf{Problem Space} & \textbf{Sensor setup} & \textbf{Size} & \textbf{Location} & \textbf{\begin{tabular}[c]{@{}c@{}}Traffic \\ condition\end{tabular}} & \textbf{License} \\ \hline
\begin{tabular}[c]{@{}c@{}}NuScenes\\ \cite{nuscenes2019}\end{tabular} & \begin{tabular}[c]{@{}c@{}}3D tracking, \\ 3D object \\ detection\end{tabular}                                         & \begin{tabular}[c]{@{}c@{}}Radar, Lidar, \\ EgoData, GPS,\\ IMU, Camera\end{tabular}    & \begin{tabular}[c]{@{}c@{}}345 GB \\ (1000 scenes, clips of 20s)\end{tabular} & \begin{tabular}[c]{@{}c@{}}Boston,\\  Singapore\end{tabular}                & Urban                                                     & CC BY-NC-SA 3.0                 \\ \hline
\begin{tabular}[c]{@{}c@{}}AMUSE\\ \cite{amuse2013}\end{tabular}                                                      & SLAM                                                                                                                 & \begin{tabular}[c]{@{}c@{}}Omnidirectional\\ camera, IMU,\\ EgoData, GPS\end{tabular}     & \begin{tabular}[c]{@{}c@{}} 1 TB \\ (7 clips)\end{tabular}         & Los Angeles                                                                 & Urban                                                     & CC BY-NC-ND 3.0                 \\ \hline
\begin{tabular}[c]{@{}c@{}}Ford\\ \cite{Pandey2011}\end{tabular}                                                      & \begin{tabular}[c]{@{}c@{}}3D tracking,\\  3D object detection\end{tabular}                                          & \begin{tabular}[c]{@{}c@{}}Omnidirectional\\ camera,  IMU,\\Lidar, GPS\end{tabular}          & 100 GB                                                                   & Michigan                                                                    & Urban                                                     & Not specified                   \\ \hline
\begin{tabular}[c]{@{}c@{}}KITTI\\ \cite{KITTI2013}\end{tabular}                                                        & \begin{tabular}[c]{@{}c@{}}3D tracking, \\ 3D object detection, \\ SLAM\end{tabular}                                 & \begin{tabular}[c]{@{}c@{}}Monocular\\ cameras, IMU\\  Lidar, GPS\end{tabular}               & 180 GB                                                                   & Karlsruhe                                                                   & \begin{tabular}[c]{@{}c@{}}Urban \\ Rural\end{tabular}    & CC BY-NC-SA 3.0                 \\ \hline
\begin{tabular}[c]{@{}c@{}}Udacity\\ \cite{Udacity2018}\end{tabular}                                    & \begin{tabular}[c]{@{}c@{}}3D tracking, \\ 3D object detection\end{tabular}                                          & \begin{tabular}[c]{@{}c@{}}Monocular \\cameras, IMU,\\Lidar, GPS,\\ EgoData\end{tabular}    & 220 GB                                                                   & Mountain View                                                               & Rural                                                     & MIT                             \\ \hline
\begin{tabular}[c]{@{}c@{}}Cityscapes\\ \cite{Cityscapes2018}\end{tabular}                                                   & \begin{tabular}[c]{@{}c@{}}Semantic \\ understanding\end{tabular}                                                    & \begin{tabular}[c]{@{}c@{}}Color stereo \\cameras                                                                       \end{tabular} & \begin{tabular}[c]{@{}c@{}} 63 GB \\ (5 clips)\end{tabular}        & \begin{tabular}[c]{@{}c@{}}Darmstadt, \\ Zurich, \\ Strasbourg\end{tabular} & Urban                                                     & CC BY-NC-SA 3.0                 \\ \hline
\begin{tabular}[c]{@{}c@{}}Oxford\\ \cite{Oxford2018}\end{tabular}                                                    & \begin{tabular}[c]{@{}c@{}}3D tracking, \\ 3D object detection,\\  SLAM\end{tabular}                                 & \begin{tabular}[c]{@{}c@{}}Stereo and \\ monocular \\ cameras, GPS\\ Lidar, IMU\end{tabular} & \begin{tabular}[c]{@{}c@{}}  23 TB \\  (133 clips)\end{tabular}       & Oxford                                                                      & \begin{tabular}[c]{@{}c@{}}Urban, \\ Highway\end{tabular} & CC BY-NC-SA 3.0                 \\ \hline
\begin{tabular}[c]{@{}c@{}}CamVid\\ \cite{CamVid2018}\end{tabular}                                                      & \begin{tabular}[c]{@{}c@{}}Object detection, \\ Segmentation\end{tabular}                                            & \begin{tabular}[c]{@{}c@{}}Monocular \\color \\ camera\end{tabular}                              & \begin{tabular}[c]{@{}c@{}} 8 GB \\ (4 clips)\end{tabular}                     & Cambridge                                                                   & Urban                                                     & N/A                             \\ \hline
\begin{tabular}[c]{@{}c@{}}Daimler\\pedestrian\\ \cite{Gavrila2013}\end{tabular} & \begin{tabular}[c]{@{}c@{}}Pedestrian detection, \\ Classification, \\ Segmentation, \\ Path prediction\end{tabular} & \begin{tabular}[c]{@{}c@{}}Stereo and\\ monocular \\ cameras\end{tabular}                        & \begin{tabular}[c]{@{}c@{}} 91 GB \\ (8 clips)\end{tabular}               & \begin{tabular}[c]{@{}c@{}}Amsterdam, \\ Beijing\end{tabular}               & Urban                                                     & N/A                             \\ \hline
\begin{tabular}[c]{@{}c@{}}Caltech\\ \cite{Dollar2009}\end{tabular}                                                     & \begin{tabular}[c]{@{}c@{}}Tracking, \\ Segmentation, \\ Object detection\end{tabular}                               & \begin{tabular}[c]{@{}c@{}}Monocular \\camera                                                                              \end{tabular} & 11 GB                                                                    & \begin{tabular}[c]{@{}c@{}}Los Angeles\\  (USA)\end{tabular}                & Urban                                                     & N/A                      \\ \hline
\end{tabular}}
\caption{Summary of datasets for training autonomous driving systems}
\label{tab:datasets}
\end{table*}

\textit{KITTI Vision Benchmark dataset (KITTI)} \cite{KITTI2013}. Provided by the Karlsruhe Institute of Technology (KIT) from Germany, this dataset fits the challenges of benchmarking stereo-vision, optical flow, 3D tracking, 3D object detection or SLAM algorithms. It is known as the most prestigious dataset in the self-driving vehicles domain. To this date it counts more than 2000 citations in the literature. The data collection vehicle is equipped with multiple high-resolution color and gray-scale stereo cameras, a Velodyne 3D LiDAR and high-precision GPS/IMU sensors. In total, it provides 6 hours of driving data collected in both rural and highway traffic scenarios around Karlsruhe. The dataset is provided under the Creative Commons Attribution-NonCommercial-ShareAlike 3.0 License.

\textit{NuScenes dataset} \cite{nuscenes2019}. Constructed by nuTonomy, this dataset contains 1000 driving scenes collected from Boston and Singapore, two known for their dense traffic and highly challenging driving situations. In order to facilitate common computer vision tasks, such as object detection and tracking, the providers annotated 25 object classes with accurate 3D bounding boxes at 2Hz over the entire dataset. Collection of vehicle data is still in progress. The final dataset will include approximately 1,4 million camera images, 400.000 Lidar sweeps, 1,3 million RADAR sweeps and 1,1 million object bounding boxes in 40.000 keyframes. The dataset is provided under the Creative Commons Attribution-NonCommercial-ShareAlike 3.0 License license.

\textit{Automotive multi-sensor dataset (AMUSE)} \cite{amuse2013}. Provided by Link\"{o}ping University of Sweden, it consists of sequences recorded in various environments from a car equipped with an omnidirectional multi-camera, height sensors, an IMU, a velocity sensor and a GPS. The API for reading these data sets is provided to the public, together with a collection of long multi-sensor and multi-camera data streams stored in the given format. The dataset is provided under the Creative Commons Attribution-NonCommercial-NoDerivs 3.0 Unsupported License.

\textit{Ford campus vision and lidar dataset (Ford)} \cite{Pandey2011}. Provided by University of Michigan, this dataset was collected using a Ford F250 pickup truck equipped with professional (Applanix POS-LV) and a consumer (Xsens MTi-G) inertial measurement units (IMU), a Velodyne Lidar scanner, two push-broom forward looking Riegl Lidars and a Point Grey Ladybug3 omnidirectional camera system. The approx. 100 GB of data was recorded around the Ford Research campus and downtown Dearborn, Michigan in 2009. The dataset is well suited to test various autonomous driving and simultaneous localization and mapping (SLAM) algorithms.

\textit{Udacity dataset} \cite{Udacity2018}. The vehicle sensor setup contains monocular color cameras, GPS and IMU sensors, as well as a Velodyne 3D Lidar. The size of the dataset is 223GB. The data is labeled and the user is provided with the corresponding steering angle that was recorded during the test runs by the human driver.

\textit{Cityscapes dataset}\cite{Cityscapes2018}. Provided by Daimler AG R\&D, Germany; Max Planck Institute for Informatics (MPI-IS), Germany, TU Darmstadt Visual Inference Group, Germany, the Cityscapes Dataset focuses on semantic understanding of urban street scenes, this being the reason for which it contains only stereo vision color images. The diversity of the images is very large: 50 cities, different seasons (spring, summer, fall), various weather conditions and different scene dynamics. There are 5000 images with fine annotations and 20000 images with coarse annotations. Two important challenges have used this dataset for benchmarking the development of algorithms for semantic segmentation \cite{Zhao2017} and instance segmentation \cite{Liu2017}.  

\textit{The Oxford dataset} \cite{Oxford2018}. Provided by Oxford University, UK, the dataset collection spanned over 1 year, resulting in over 1000 km of recorded driving with almost 20 million images collected from 6 cameras mounted to the vehicle, along with LIDAR, GPS and INS ground truth. Data was collected in all weather conditions, including heavy rain, night, direct sunlight and snow. One of the particularities of this dataset is that the vehicle frequently drove the same route over the period of a year to enable researchers to investigate long-term localization and mapping for autonomous vehicles in real-world, dynamic urban environments.

\textit{The Cambridge-driving Labeled Video Dataset (CamVid)} \cite{CamVid2018}. Provided by the University of Cambridge, UK, it is one of the most cited dataset from the literature and the first released publicly, containing a collection of videos with object class semantic labels, along with metadata annotations. The database provides ground truth labels that associate each pixel with one of 32 semantic classes. The sensor setup is based on only one monocular camera mounted on the dashboard of the vehicle. The complexity of the scenes is quite low, the vehicle being driven only in urban areas with relatively low traffic and good weather conditions. 

\textit{The Daimler pedestrian benchmark dataset} \cite{Gavrila2013}. Provided by Daimler AG R\&D and University of Amsterdam, this dataset fits the topics of pedestrian detection, classification, segmentation and path prediction. Pedestrian data is observed from a traffic vehicle by using only on-board mono and stereo cameras. It is the first dataset with contains pedestrians. Recently, the dataset was extended with cyclist video samples captured with the same setup \cite{Flohr2016}. 

\textit{Caltech pedestrian detection dataset (Caltech)} \cite{Dollar2009}. Provided by California Institute of Technology, US, the dataset contains richly annotated videos, recorded from a moving vehicle, with challenging images of low resolution and frequently occluded people. There are approx. 10 hours of driving scenarios cumulating about 250.000 frames with a total of 350 thousand bounding boxes and 2.300 unique pedestrians annotations. The annotations include both temporal correspondences between bounding boxes and detailed occlusion labels.

Given the variety and complexity of the available databases, choosing one or more to develop and test an autonomous driving component may be difficult. As it can be observed, the sensor setup varies among all the available databases. For localization and vehicle motion, the Lidar and GPS/IMU sensors are necessary, with the most popular Lidar sensors used being Velodyne \cite{Velodyne2018} and Sick \cite{Sick2018}. Data recorded from a radar sensor is present only in the NuScenes dataset. The radar manufacturers adopt proprietary data formats which are not public. Almost all available datasets include images captured from a video camera, while there is a balance use of monocular and stereo cameras mainly configured to capture gray-scale images. AMUSE and Ford databases are the only ones that use omnidirectional cameras.

Besides raw recorded data, the datasets usually contain miscellaneous files such as annotations, calibration files, labels, etc. In order to cope with this files, the dataset provider must offer tools and software that enable the user to read and post-process the data. Splitting of the datasets is also an important factor to consider, because some of the datasets (e.g. Caltech, Daimler, Cityscapes) already provide pre-processed data that is classified in different sets: training, testing and validation. This enables benchmarking of desired algorithms against similar approaches to be consistent.

Another aspect to consider is the \textit{license} type. The most commonly used license is Creative Commons Attribution-NonCommercial-ShareAlike 3.0. It allows the user to copy and redistribute in any medium or format and also to remix, transform, and build upon the material. KITTI and NuScenes databases are examples of such distribution license. The Oxford database uses a Creative Commons Attribution-Noncommercial 4.0. which, compared with the first license type, does not force the user to distribute his contributions under the same license as the database. Opposite to that, the AMUSE database is licensed under Creative Commons Attribution-Noncommercial-noDerivs 3.0 which makes the database illegal to distribute if modification of the material are made. 

With very few exceptions, the datasets are collected from a single city, which is usually around university campuses or company locations in Europe, the US, or Asia. Germany is the most active country for driving recording vehicles. Unfortunately, all available datasets together cover a very small portion of the world map. One reason for this is the memory size of the data which is in direct relation with the sensor setup and the quality. For example, the Ford dataset takes around 30 GB for each driven kilometer, which means that covering an entire city will take hundreds of TeraBytes of driving data. The majority of the available datasets consider sunny, daylight and urban conditions, these being ideal operating conditions for autonomous driving systems.


\section{Computational Hardware and Deployment}
\label{sec:hardware}

Deploying deep learning algorithms on target edge devices is not a trivial task. The main limitations when it comes to vehicles are the price, performance issues and power consumption. Therefore, embedded platforms are becoming essential for integration of AI algorithms inside vehicles due to their portability, versatility, and energy efficiency. 

The market leader in providing hardware solutions for deploying deep learning algorithms inside autonomous cars is NVIDIA\textsuperscript{\textregistered}. DRIVE PX \cite{DrivePX} is an AI car computer which was designed to enable the auto-makers to focus directly on the software for autonomous vehicles.

The newest version of DrivePX architecture is based on two Tegra X2~\cite{TegraX2} systems on a chip (SoCs). Each SoC contains two Denve~\cite{Denver} cores, 4 ARM A57 cores and a graphical computeing unit (GPU) from the Pascal \cite{Pascal} generation. NVIDIA\textsuperscript{\textregistered} DRIVE PX is capable to perform real-time environment perception, path planning and localization. It combines deep learning, sensor fusion and surround vision to improve the driving experience. 

Introduced in September 2018, NVIDIA\textsuperscript{\textregistered} DRIVE AGX developer kit platform was presented as the world's most advanced self-driving car platform~\cite{DriveAGX}, being based on the Volta technology~\cite{Volta}. It is available in two different configurations, namely DRIVE AGX Xavier and DRIVE AGX Pegasus. 

DRIVE AGX Xavier is a scalable open platform that can serve as an AI brain for self driving vehicles, and is an energy-efficient computing platform, with 30 trillion operations per second, while meeting automotive standards like the ISO 26262 functional safety specification. NVIDIA\textsuperscript{\textregistered} DRIVE AGX Pegasus improves the performance with an architecture which is built on two NVIDIA\textsuperscript{\textregistered} Xavier processors and two state of the art TensorCore GPUs.

A hardware platform used by the car makers for Advanced Driver Assistance Systems (ADAS) is the R-Car V3H system-on-chip (SoC) platform from Renesas Autonomy~\cite{Renesas}. This SoC provides the possibility to implement high performance computer vision with low power consumption. R-Car V3H is optimized for applications that involve the usage of stereo cameras, containing dedicated hardware for convolutional neural networks, dense optical flow, stereo-vision, and object classification. The hardware features four 1.0 GHz Arm Cortex-A53 MPCore cores, which makes R-Car V3H a suitable hardware platform which can be used to deploy trained inference engines for solving specific deep learning tasks inside the automotive domain.

Renesas also provides a similar SoC, called R-Car H3~\cite{RenesasH3} which delivers improved computing capabilities and compliance with functional safety standards. Equipped with new CPU cores (Arm Cortex-A57), it can be used as an embedded platform for deploying various deep learning algorithms, compared with R-Car V3H, which is only optimized for CNNs.

\null

A Field-Programmable Gate Array (FPGA) is another viable solution, showing great improvements in both performance and power consumption in deep learning applications. The suitability of the FPGAs for running deep learning algorithms can be analyzed from four major perspectives: efficiency and power, raw computing power, flexibility and functional safety. Our study is based on the research published by Intel~\cite{Nurvitadhi2017}, Microsoft~\cite{accelerating-deep-convolutional-hardware} and UCLA~\cite{Cong2018}.

By reducing the latency in deep learning applications, FPGAs provide additional raw computing power. The memory bottlenecks, associated with external memory accesses, are reduced or even eliminated by the high amount of chip cache memory. In  addition, FPGAs have the advantages of supporting a full range of data types, together with custom user-defined types.  

FPGAs are optimized when it comes to efficiency and power consumption. The studies presented by manufacturers like Microsoft and Xilinx show that GPUs can consume upon ten times more power than FPGAs when processing algorithms with the same computation complexity, demonstrating that FPGAs can be a much more suitable solution for deep learning applications in the automotive field. 

In terms of flexibility, FPGAs are built with multiple architectures, which are a mix of hardware programmable resources, digital signal processors and Processor Block RAM (BRAM) components. This architecture flexibility is suitable for deep and sparse neural networks, which are the state of the art for the current machine learning applications. Another advantage is the possibility of connecting to various input and output peripheral devices like sensors, network elements and storage devices.

In the automotive field, functional safety is one of the most important challenges. FPGAs have been designed to meet the safety requirements for a wide range of applications, including ADAS. When compared to GPUs, which were originally built for graphics and high-performance computing systems, where functional safety is not necessary, FPGAs provide a significant advantage in developing driver assistance systems.

\section{Discussion and Conclusions}
\label{sec:conclusion}

We have identified seven major areas that form open challenges in the field of autonomous driving. We believe that Deep Learning and Artificial Intelligence will play a key role in overcoming these challenges:

\textit{Perception}: In order for an autonomous car to safely navigate the driving scene, it must be able to understand its surroundings. Deep learning is the main technology behind a large number of perception systems. Although great progress has been reported with respect to accuracy in object detection and recognition~\cite{zhao2018object}, current systems are mainly designed to calculate 2D or 3D bounding boxes for a couple of trained object classes, or to provide a segmented image of the driving environment. Future methods for perception should focus on increasing the levels of recognized details, making it possible to perceive and track more objects in real-time. Furthermore, additional work is required for bridging the gap between image- and LiDAR-based 3D perception~\cite{wang2019pseudo}, enabling the computer vision community to close the current debate on camera vs. LiDAR as main perception sensors.

\textit{Short- to middle-term reasoning}: Additional to a robust and accurate perception system, an autonomous vehicle should be able to reason its driving behavior over a short (milliseconds) to middle (seconds to minutes) time horizon~\cite{Pendleton2017}. AI and deep learning are promising tools that can be used for the high- and low-level path path planning required for navigating the miriad of driving scenarios. Currently, the largest portion of papers in deep learning for self-driving cars are focused mainly on perception and End2End learning~\cite{shalevshwartz2016safe,Zhang_2016}. Over the next period, we expect deep learning to play a significant role in the area of local trajectory estimation and planning. We consider long-term reasoning as solved, as provided by navigation systems. These are standard methods for selecting a route through the road network, from the car's current position to destination~\cite{Pendleton2017}.

\textit{Availability of training data}: "Data is the new oil" became lately one of the most popular quote in the automotive industry. The effectiveness of deep learning systems is directly tied to the availability of training data. As a rule of thumb, current deep learning methods are also evaluated based on the quality of training data~\cite{Janai2017}. The better the quality of the data is, the higher the accuracy of the algorithm. The daily data recorded by an autonomous vehicle is on the order of petabytes. This poses challenges on the parallelization of the training procedure, as well as on the storage infrastructure. Simulation environments have been used in the last couple of years for bridging the gap between scarce data and the deep learning's hunger for training examples. There is still a gap to be filled between the accuracy of a simulated world and real-world driving.

\textit{Learning corner cases}: Most driving scenarios are considered solvable with classical methodologies. However, the remaining unsolved scenarios are corner cases which, until now, required the reasoning and intelligence of a human driver. In order to overcome corner cases, the generalization power of deep learning algorithms should be increased. Generalization in deep learning is of special importance in learning hazardous situations that can lead to accidents, especially due to the fact that training data for such corner cases is scarce. This implies also the design of \textit{one-shot} and \textit{low-shot} learning methods, that can be trained a reduced number of training examples.

\textit{Learning-based control methods}: Classical controllers make use of an a-priori model composed of fixed parameters. In a complex case, such as autonomous driving, these controllers cannot anticipate all driving situations. The effectiveness of deep learning components to adapt based on past experiences can also be used to learn the parameters of the car's control system, thus better approximating the underlaying true system model~\cite{Ostafew2016,ostafew-ijrr16}.

\textit{Functional safety}: The usage of deep learning in safety-critical systems is still an open debate, efforts being made to bring the computational intelligence and functional safety communities closer to each other. Current safety standards, such as the ISO 26262, do not accommodate machine learning software~\cite{Salay2017}. Although new data-driven design methodologies have been proposed, there are still opened issues on the explainability, stability, or classification robustness of deep neural networks.

\textit{Real-time computing and communication}: Finally, real-time requirements have to be fulfilled for processing the large amounts of data gathered from the car's sensors suite, as well as for updating the parameters of deep learning systems over high-speed communication lines~\cite{Nurvitadhi2017}. These real-time constraints can be backed up by advances in semiconductor chips dedicated for self-driving cars, as well as by the rise of 5G communication networks.

\subsection{Final Notes}

Autonomous vehicle technology has seen a rapid progress in the past decade, especially due to advances in the area of artificial intelligence and deep learning. Current AI methodologies are nowadays either used or taken into consideration when designing different components for a self-driving car. Deep learning approaches have influenced not only the design of traditional perception-planning-action pipelines, but have also enabled End2End learning systems, able do directly map sensory information to steering commands.

Driverless cars are complex systems which have to safely drive passengers or cargo from a starting location to destination. Several challenges are encountered with the advent of AI based autonomous vehicles deployment on public roads. A major challenge is the difficulty in proving the functional safety of these vehicle, given the current formalism and explainability of neural networks. On top of this, deep learning systems rely on large training databases and require extensive computational hardware.

This paper has provided a survey on deep learning technologies used in autonomous driving. The survey of performance and computational requirements serves as a reference for system level design of AI based self-driving vehicles.

\subsection*{Acknowledgment}

The authors would like to thank Elektrobit Automotive for the infrastructure and research support.


\bibliographystyle{IEEEtran}
\bibliography{references}

\end{document}